\documentclass[10pt,journal,compsoc]{IEEEtran_tpami}
\usepackage{color,xcolor}
\usepackage{graphicx}
\usepackage{algorithm}
\usepackage{algorithmicx}
\usepackage[noend]{algpseudocode}

\usepackage{amssymb}
\usepackage{amsmath}
\usepackage{subfigure}
\usepackage{booktabs}
\usepackage{multirow}
\usepackage{stackengine}
\usepackage{amssymb}
\usepackage{pifont}
\usepackage{caption}

\newcommand{\m}[1]{\mathbf{#1}}
\newcommand{\ms}[1]{\boldsymbol{#1}}
\newcommand{\tabincell}[2]{\begin{tabular}{@{}#1@{}}#2\end{tabular}}
\newcommand{\cmark}{\ding{51}}%
\newcommand{\xmark}{\ding{55}}%
\newcommand{\TODO}[1]{}
\newcommand*\rot{\rotatebox{90}}

\usepackage{diagbox}
\usepackage[utf8]{inputenc}
\usepackage[english]{babel}
\usepackage{hyperref}

\ifCLASSINFOpdf
\else
\fi
\begin{document}
%
\title{Towards Efficient Post-training Quantization of Pre-trained Language Models}
%
%
%

\author{Haoli~Bai,~\IEEEmembership{Student Member,~IEEE,}
        Lu~Hou, 
        Lifeng~Shang,
        Xin~Jiang,\\
        Irwin~King,~\IEEEmembership{Fellow,~IEEE,}
        Michael~R.~Lyu,~\IEEEmembership{Fellow,~IEEE}
\thanks{H. Bai, I. King and M. Lyu are with the Department
of Computer Science and Engineering, The Chinese University of Hong Kong, Hong Kong, 
e-mail: \{hlbai, king, lyu\}@cse.cuhk.edu.hk.}
\thanks{L. Hou, L. Shang and X. Jiang are with Huawei Noah's Ark Lab, e-mail: \{houlu3, shang.lifeng, jiang.xin\}@huawei.com}.
}

\IEEEtitleabstractindextext{
\begin{abstract}
Network quantization has gained increasing attention with the rapid growth of large pre-trained language models~(PLMs).
However, most existing quantization methods for PLMs follow quantization-aware training~(QAT) that requires 
end-to-end training with full access to the entire dataset. 
Therefore, they suffer from slow training, large memory overhead, and data security issues.
In this paper, we study post-training quantization~(PTQ) of PLMs, and propose module-wise quantization error minimization~(MREM), an efficient solution to mitigate these issues.
By partitioning the PLM into multiple modules, we minimize the reconstruction error incurred by quantization for each module.
In addition, we design a new model parallel training strategy such that each module can be trained locally on separate computing devices without waiting for preceding modules, which brings nearly the theoretical training speed-up (e.g., $4\times$ on $4$ GPUs).
Experiments on GLUE and SQuAD benchmarks show that our proposed PTQ solution not only performs close to QAT, but also enjoys significant reductions in training time, memory overhead, and data consumption.
\end{abstract}
}


\maketitle

\IEEEdisplaynontitleabstractindextext

%
\IEEEpeerreviewmaketitle


\section{Introduction}
\IEEEPARstart{L}{arge} pre-trained language models~(PLMs) have achieved remarkable success in various natural language processing tasks~\cite{vaswani2017attention,devlin2019bert,brown2020language}.
However, the increasing size and computation overhead also make it prohibitive to deploy these PLMs on resource-constrained devices. 
To obtain compact PLMs, various model compression methods have been proposed, such as  pruning~\cite{michel2019sixteen,fan2019reducing}, 
knowledge distillation~\cite{sanh2019distilbert,sun2019patient,jiao2020tinybert}, weight-sharing~\cite{dehghani2019universal,lan2020albert,wang2020revisiting,huang2021ghostbert},
dynamic computation with adaptive depth or width~\cite{hou2020dynabert,xin2020deebert,zhou2020bert}, 
and quantization~\cite{zafrir2019q8bert,shen2020qbert,zadeh2020gobo,zhang2020ternarybert,bai2021binarybert}.

 
Among these methods, network quantization enjoys the reduction of both model size and computation overhead without modifying the network architecture.
However, despite its remarkable performance, prior methods mostly follow quantization-aware training~(QAT) and thus suffer from multiple challenges: 
1) QAT generally requires the same order of  training iterations with the original full-precision training~\cite{shen2020qbert,zhang2020ternarybert,bai2021binarybert}, which can be slow to obtain the quantized model;
2) QAT usually adopts end-to-end training by back-propagation, while it can be challenging to load the entire PLM into memory on resource-limited devices.
Moreover, recent QAT efforts combine knowledge distillation to enhance the performance~\cite{zhang2020ternarybert,bai2021binarybert}, and thus further increase the memory overhead due to the teacher model;
3) QAT needs full access to the entire training set, which may lead to data security issues when exposing them to third-party organizations for the quantization service.

 \begin{figure}[t]
 	\centering
    \includegraphics[width=0.46\textwidth]{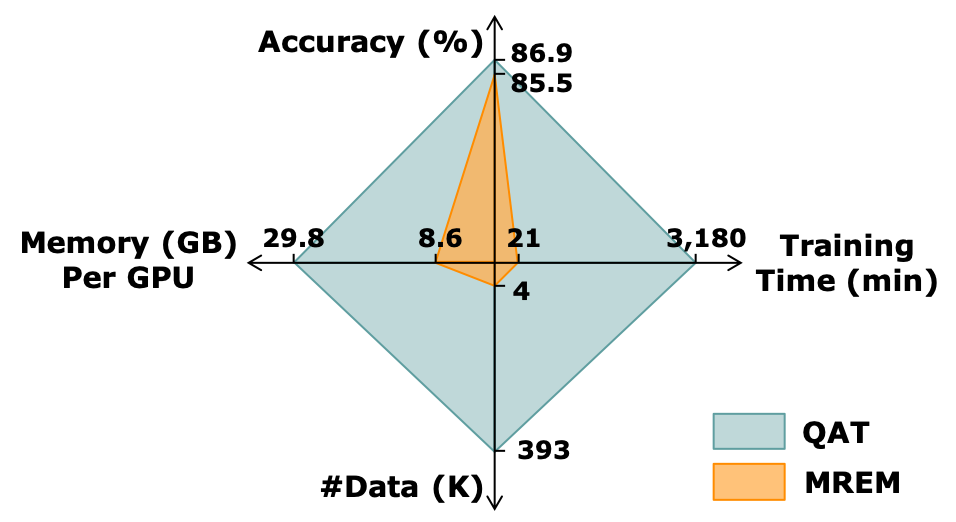}
 	\caption{An illustrative comparison between our parallel post-training quantization method (MREM) and QAT on four dimensions of the quantization pipeline: accuracy, training time, memory overhead, and data consumption. The results are based on a quantized BERT-large model with 4-bit weights and 8-bit activations over the MNLI dataset. Best viewed in color.
 	}
 	\label{fig:intro}
 \end{figure}

Given the above challenges, post-training quantization~(PTQ) serves as an appealing alternative.
In contrast to QAT, PTQ is efficient in both training time, memory overhead and data consumption. Instead of the full training set, it is common in PTQ to adopt only a small portion of training data 
to minimize the reconstruction error incurred by quantization~\cite{nagel2019data,nahshan2019loss,nagel2020up,hubara2020improving}. 
This can be done by calibrating the batch normalization  statistics~\cite{nagel2019data} or step sizes~\cite{nahshan2019loss} in quantization functions in a layer-wise manner.
The layer-wise objective is also more sample-efficient~\cite{zhou2020go} and memory-saving compared with end-to-end training in QAT.
Despite the success of prior PTQ solutions on convolutional neural networks~(CNNs), we show that it is non-trivial to directly apply them to PLMs such as BERT~\cite{devlin2019bert}. Different from CNNs, there are multiple linear layers coupled together within the multi-head self-attention and feed-forward network of the transformer architecture. Therefore, layer-wise training ignores the underlying correlation among layers and thus leads to poor performance.




In this paper, we aim at improving the performance of post-training quantization for PLM, while simultaneously maintaining its efficiency w.r.t training time, memory overhead and data accessibility.
Firstly, we propose \textbf{\textit{module-wise reconstruction error minimization}}~(MREM) to incorporate more layer-wise correlation.
By partitioning the PLM into multiple modules, each module consists of multiple Transformer layers for joint optimization.
Meanwhile, the module size can be flexibly adjusted depending on the memory constraints, achieving an appropriate trade-off between layer-wise correlation and memory overhead.
While similar block-wise objectives are previously considered in~\cite{li2021brecq}, they require to compute second-order Hessian matrices for optimization, which can be computationally prohibitive for large PLMs.
Secondly, we design a new \textbf{\textit{model parallel strategy}} to further accelerate the process of MREM.
By allocating each module to an individual computing device, all modules can perform local training in parallel, achieving nearly the theoretical speed-up (e.g., $4\times$ on $4$ GPUs).
Thirdly, we develop \textbf{\textit{annealed teacher forcing}} for the parallel training. 
We find that the naive parallel training suffers from the propagation of reconstruction error, 
since each quantized module passes the error to its successors before it is fully converged.
Inspired by~\cite{williams1989learning}, we use the full-precision module to provide clean signals to the next quantized module. This breaks the reconstruction error propagation and improves the performance of quantized PLMs.


Empirical results on the GLUE and SQuAD benchmarks show that our proposed MREM not only significantly improves the performance for post-training quantization,
but also enjoys advantages of faster training, less memory overhead, and improved data security over QAT.
For instance, as is shown in Figure~\ref{fig:intro}, the BERT-large model trained by parallel MREM can achieve $85.5\%$ accuracy based on only $4$K training samples. 
Moreover, it consumes merely one-third of memory per GPU and is more than $150\times$ faster than previous QAT training.

We summarize the contributions of this paper as follows:
\begin{itemize}
    \item We study the post-training quantization of PLMs, and propose module-wise reconstruction error minimization~(MREM), a fast, memory-saving, and data-efficient approach to improve the quantized PLMs.
    \item We design a new model parallel strategy based on MREM to accelerate post-training quantization with theoretical speed-up for distributed training.
    \item The parallel MREM can be combined with annealed teacher forcing to alleviate the propagation of reconstruction error and boost the performance.
    \item We conduct extensive experiments on both GLUE and SQuAD benchmarks to verify the advantages of our approach w.r.t. multiple aspects of the quantization pipeline. We also provide detailed discussions on other important factors of our approach.
\end{itemize}

The rest of this paper is organized as follows. We summarize the related work in Section~\ref{sec:related}. In Section~\ref{sec:motivation}, we review the necessary backgrounds and explain the motivation for this research. In Section~\ref{sec:method}, we propose our PTQ solution together with the parallel training technique. The experiments are present in Section~\ref{sec:exp}. Finally, we conclude this work in Section~\ref{sec:conclusion}.




\section{Related Work}
\label{sec:related}

In this section, we review the literature related to our research, including both network compression and parallel training for pre-trained language models. 

\subsection{Network Compression for Pre-trained Language Models}

As the research focus in this paper, network quantization replaces the original full-precision parameters and activations with low-bit representations~\cite{courbariaux2015binaryconnect, li2016ternary,hou2017loss,hou2018loss,esser2019learned,li2020rtn,zhuang2021effective,young2021transform}. 
To apply network quantization on PLMs, Q8BERT~\cite{zafrir2019q8bert} convert both parameters and activations with 8-bit representations, and finds that there is negligible accuracy drop on natural language understanding tasks.
Q-BERT~\cite{shen2020qbert} exploit the hessian matrix of loss curvature to determine the best layer-wise quantization bit-width, which achieves a higher compression rate. Additionally, TernaryBERT~\cite{zhang2020ternarybert} propose to ternarize the parameters with 2-bit representations together with two-stage knowledge distillation~\cite{jiao2020tinybert}.
Recently, BinaryBERT~\cite{bai2021binarybert} binarize the model parameters based by splitting from ternarized models, which further reduces the quantization bit-width. 
Despite the promising performance of these quantization approaches, they mostly follow quantization-aware training that requires heavy fine-tuning, which can be prohibitive given constraints on the training time, memory size, and data accessibility. 
In this work, we shall follow the post-training quantization~\cite{nagel2019data,nahshan2019loss,zhao2019improving,nagel2020up,li2021brecq,hubara2020improving}, the other way to improve the quantized PLMs given limited training resources. 
More details on quantization will be discussed in Section~\ref{sec:motivation}.

Aside from quantization, there are also several other popular techniques to compress PLMs.
Pruning removes unimportant parameters or connections in a well-trained model~\cite{he2017channelprunning,luo2018thinet,wen2019structured,wang2020bayesian,liu2021discrimination}, and is widely explored in PLMs~\cite{gordon2020compressing,wang2020rethinking}.
A direct way is to remove connections with small magnitudes during the pre-training and adds them back when necessary in downstream tasks~\cite{gordon2020compressing}.
Structured pruning can also be applied to directly reduce the width of BERT~\cite{McCarley2019q}, which is more friendly for practical inference acceleration.
In~\cite{michel2019sixteen}, it is shown that attention heads can also be pruned without hurting the representation of PLMs.
Furthermore, a comprehensive study is provided in~\cite{wang2020rethinking} to investigate the pruning sensitivity of different parts of the Transformer model.
There are also efforts on dropping the entire layers of transformer models~\cite{fan2019reducing,xin2020deebert}.

Knowledge distillation~\cite{hinton2015distilling,romero2014fitnets,wang2021distilling,zhang2021self,liu2020structured} is another successful tool to 
design efficient PLMs, by distilling knowledge from a large teacher model to a smaller student model.
Knowledge distillation is first applied to PLMs by minimizing the soft cross-entropy between output logits~\cite{sanh2019distilbert}.
Aside from this,
recent efforts show that it is also helpful to minimize the mean square error of hidden representations~\cite{jiao2020tinybert,sun2019patient}.
While knowledge distillation is promising in performance, 
memory consumption is the main concern as the teacher model 
itself or the size of its pre-computed representation for distillation is generally large.
Post-training quantization is also closely related to knowledge distillation~\cite{nahshan2019loss,nagel2019data,nagel2020up,li2021brecq}, when the full-precision model acts as the teacher to provide layer-wise supervision signals to the quantized student model.

An orthogonal direction is to apply neural architecture search~(NAS) for efficient PLMs structures. AdaBERT~\cite{chen2020adabert} adopt the differentiable search~\cite{liu2018darts} to automatically compress BERT into task-specific architectures. 
To obtain task-agnostic architectures, NAS can also be applied during the pre-training stage~\cite{so2019evolved,xu2021bert}.
Recently, one-shot NAS is also developed to search tiny PLMs~\cite{yin2021autotinybert}.
Despite the promising performance of these approaches, the algorithmic efficiency~\cite{bender2019understanding,wang2020revisiting} is a major concern for NAS-based PLMs.

\subsection{Parallel Training for Pre-trained Language Models}
Parallel training is also a popular topic in training large pre-trained language models, where pipeline model parallelism~\cite{huang2018gpipe,narayanan2019pipedream,tarnawski2020efficient,park2020hetpipe,fan2021dapple} is closely related to our proposed parallel training strategy. 
Specifically, by partitioning the model into multiple modules, pipeline parallelism similarly puts each module on individual computing devices. 
However, pipeline parallelism has a high computational cost among different workers to transmit the intermediate tensors in the forward and backward pass.
GPipe~\cite{huang2018gpipe} adopts mini-batches to reduce the bubble time, but still suffers from limited speed-up as will be discussed in Section~\ref{sec:compare_with_gpipe}. PipeDream~\cite{narayanan2019pipedream} optimizes the module partition to minimize the communication cost, but the resulting strategy still highly depends on the model architecture.
Our parallel training strategy, on the other hand, does not require real-time communication among workers and thus brings the theoretical speed-up. While the solution can be suboptimal, we  propose a novel teacher-forcing mechanism to alleviate the problem and the resultant performance is already reasonably good for post-training quantization.


Aside from pipeline model parallelism, there are several other dimensions for parallel training.
Data-parallelism~\cite{dean2008mapreduce,li2014communication} is the most widely used solution in modern deep learning frameworks, when the training data are partitioned over multiple workers.
Op-level model parallelism~\cite{jia2018exploring,shoeybi2019megatron,song2020accpar} slices the parameters over multiple works to compute local results and concatenate them together afterwards, which has a high communication cost.
Optimizer model parallelism~\cite{rajbhandari2020zero} is capable of partitioning parameters, gradients, and optimizer states into each computing device, achieving linear memory reduction with a number of workers. 
It would be promising to combine all these approaches with our parallel strategy, enabling the post-training quantization on gigantic pre-trained language models such as GPT-3~\cite{ramesh2021zero} and PanGu-$\alpha$~\cite{zeng2021pangu}.

\section{Motivation}
\label{sec:motivation}

\begin{figure*}[t]
	\subfigure[Training Time.]{
	    \includegraphics[width=0.24\textwidth]{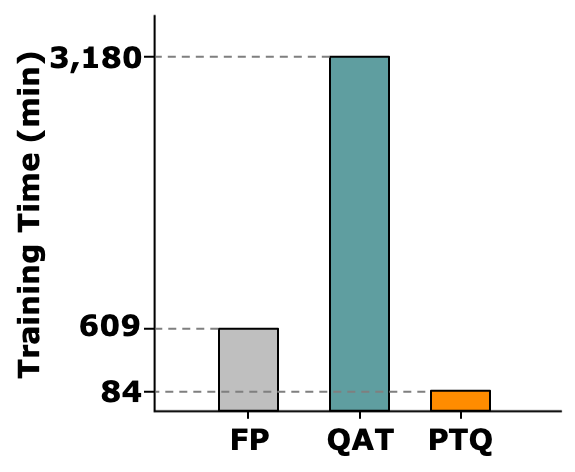}
	    \label{fig:train_time}
	}
	\subfigure[Memory Overhead.]{
	    \includegraphics[width=0.224\textwidth]{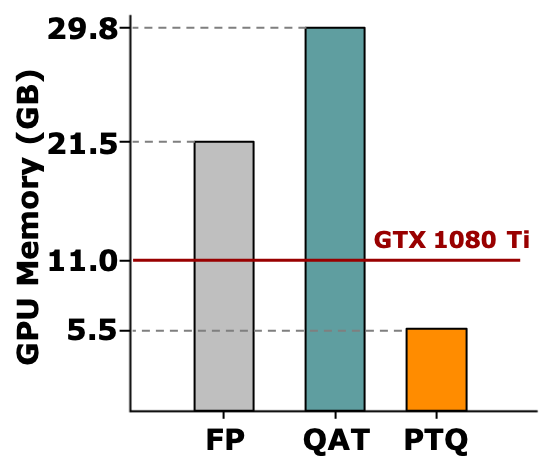}
	    \label{fig:memory}
	}
	\subfigure[Data Accessibility.]{
	    \includegraphics[width=0.224\textwidth]{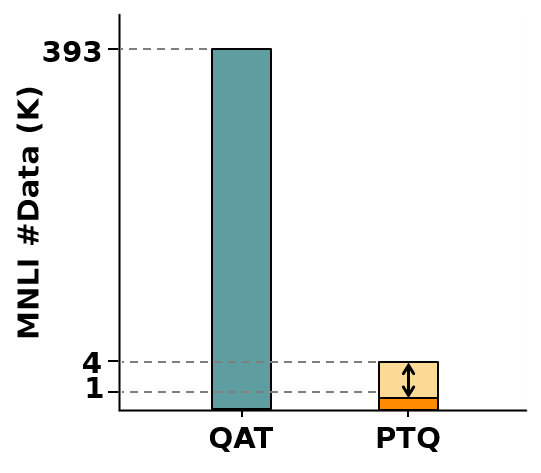}
	    \label{fig:data_num}
	}
	\subfigure[Performance.]{
	    \includegraphics[width=0.23\textwidth]{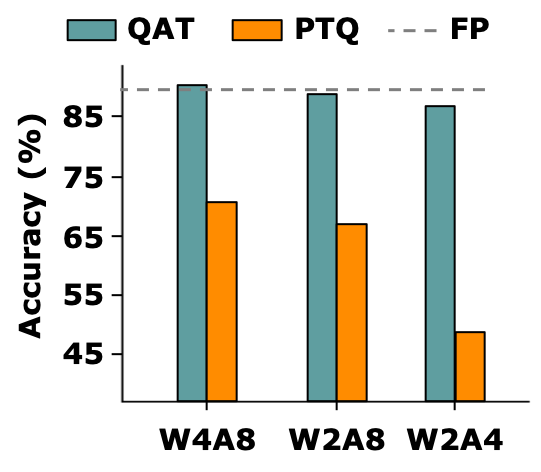}
	    \label{fig:direct_ptq}
	}
	\caption{Comparison between QAT and REM-based PTQ over four dimensions. We use a BERT-large model over MNLI dataset for illustration. 
	The full-precision~(FP) fine-tuning is also included as a baseline. 
	We follow the procedure in ~\cite{zhang2020ternarybert} for QAT, and REM in Equation~\eqref{eq:ptq_objective} for PTQ. The training time and memory in (a) and (b) are measured by 4-bit weights and 8-bit activations (i.e., W4A8) on an NVIDIA V100 GPU.}
    \label{fig:prelim}
\end{figure*}

In this section, we show that it is important yet challenging to conduct post-training quantization of PLMs. Before diving into details, we first review the necessary backgrounds for network quantization.


\subsection{Quantization Background}
\label{sec:background}
Network quantization replaces the original full-precision weight or activation
$\m x \in \mathbb{R}^{m\times n}$ with its lower-bit counterpart $\hat{\m x}$.
Denoting $s\in \mathbb{R}^{+}$ as the step size, the $b$-bit symmetric uniform quantization function $\mathcal{Q}_b(\cdot)$ can be written as
\begin{equation}
\label{eq:quantize}
    \hat{\m x} = \mathcal{Q}_b(\m x) =  s \cdot \Pi_{\Omega(b)} (\m x / s),
\end{equation}
where $\Omega(b)= \{-2^{b-1}+1, ..., 0, ..., 2^{b-1}-1\}$ is the set of $b$-bit integers, and
$\Pi(\cdot)$ is the projection function that maps $\m x / s$ to its closest integer.

In the context of 
quantization of Transformer-based PLMs, 
we follow the default setting in previous works~\cite{zafrir2019q8bert,zhang2020ternarybert,bai2021binarybert}: we quantize both the network weights and activations in each matrix multiplication. \
We use 
symmetric uniform quantization for weights, embeddings, and activations, except activations after the self-attention and GeLU function.
For these two activations, we adopt asymmetric quantization since their elements are mostly positive. We skip the quantization for all layer-normalization layers, skip connections, biases and the last classification head due to limited computation overhead or large performance drop.

In the following, we introduce two common branches in the quantization literature: quantization-aware training and post-training quantization. 

\subsubsection{Quantization-aware Training~(QAT)} 
Quantization-aware training resembles normal training, but performs the forward propagation with the quantized network.
Thus it is also time-consuming to iterate over the entire training set $\mathcal{D}$.
Typical training objective can be either the cross-entropy loss between the prediction and ground-truth labels for classification tasks~\cite{zafrir2019q8bert}, 
or the distillation objective between the quantized model and a full-precision teacher model~\cite{zhang2020ternarybert}.
As the quantization function $\mathcal{Q}_b(\cdot)$ is non-differentiable, straight-through estimator~\cite{courbariaux2015binaryconnect} is usually adopted to allow the gradient back propagation through these discrete operations.



\subsubsection{Post-training Quantization~(PTQ)} 
Unlike QAT, post-training quantization seeks to recover the performance degradation without intensive training over the entire training set $\mathcal{D}$.
One line of PTQ research quantizes the network purely without using any training data, but removes outliers in the full-precision parameters.
This can be achieved by splitting an outlier neuron with a large magnitude into two parts~\cite{zhao2019improving}, where the magnitude can be halved. Alternatively, one can scale down outlier magnitude and multiply it back in subsequent layers, a.k.a. weight equalization in ~\cite{nagel2019data}. 
Another solution is to treat the outliers and normal values in the distribution separately, by keeping two sets of quantization parameters~\cite{fang2020post,zadeh2020gobo}.





Another line of PTQ research~\cite{nahshan2019loss,wang2020towards,nagel2020up,hubara2020improving}
aims at reconstruction error minimization~(REM) using a very slight portion of unlabeled data~(a.k.a. calibration set) $\tilde{\mathcal{D}}\subseteq \mathcal{D}$ from the original training set. 
Compared with training-free PTQ approaches, such an approach is able to significantly improve the performance of the quantized network.
REM can be achieved by minimizing the distance between the output of multiplication between the quantized and the full-precision counterpart as follows:
\begin{align}
\label{eq:ptq_objective}
    & \min_{\m w, \m s} \|\hat{\m w}^{\top} \hat{\m a} - \m w^{\top} \m a\|^2, \\
    & \mathrm{s.t.}\ \hat{\m w} = \mathcal{Q}_{b_w}(\m w), \quad \hat{\m a} = \mathcal{Q}_{b_a}(\m a), \nonumber
\end{align}
where $\m w$ and $\m a$ are full-precision weights and activations, $\hat{\m w}$ and $\hat{\m a}$ are their quantized representations with $b_w$ and $b_a$ bit-widths, and $\m s$ denotes all step-sizes involved for quantization. REM is usually conducted in a greedy manner. It proceeds to the matrix multiplication only after the training of previous ones. Meanwhile, Equation~\eqref{eq:ptq_objective} can be solved quickly with the calibration set $\tilde{\mathcal{D}}$. 
Recent work~\cite{zhou2020go} also theoretically shows that such greedy objective is more sample-efficient compared with conventional end-to-end training.
In this paper, we shall extend REM-based  post-training quantization given its past success. 






\subsection{Why Post-training Quantization?}
\label{sec:why_ptq}

In this section, we discuss the difference between REM-based PTQ and QAT along four dimensions of a quantization pipeline:
1) training time; 2) memory overhead; 3) data accessibility and 4) performance. 
According to Figure~\ref{fig:prelim}, we summarize the findings in the following paragraphs.



\subsubsection{Training Time} 
As QAT iterates over the full training set $\mathcal{D}$ for multiple epochs, it is much more time-consuming than PTQ.
Note that recent QAT methods~\cite{zhang2020ternarybert,bai2021binarybert} further combine two-stage knowledge distillation~\cite{jiao2020tinybert}, which even prolongs the training compared with the full-precision~(FP) fine-tuning.
As shown in Figure~\ref{fig:train_time}, QAT can take nearly four times longer than FP.

\subsubsection{Memory Overhead} The increasing size of recent large PLMs makes it prohibited to conduct QAT on memory-limited computing resources. From Figure~\ref{fig:memory}, QAT~\cite{zhang2020ternarybert} even consumes $8.3$ GB more memory than FP when combined with knowledge distillation to store the full-precision teacher model. 
On the other hand, PTQ only caches intermediate results during the layer-wise REM in Equation~\eqref{eq:ptq_objective}, which can be fed into a single GTX 1080 Ti.
Therefore, PTQ is also applicable on memory-limited computing devices.

\subsubsection{Data Accessibility} The quantization service can be usually offered by some third-party organizations, where data security is always of high priority. 
As QAT requires access to the entire training set, it inevitably increases the risk of data exposure.
PTQ, on the other hand, needs only a small amount of calibration data $\tilde{\mathcal{D}}\subseteq\mathcal{D}$, and can be easily constructed by randomly sampling $1$K $\sim 4$K instances from $\mathcal{D}$, as shown in Figure~\ref{fig:data_num}. Therefore, most original training instances are kept untouched and data security can be largely guaranteed.

\subsubsection{Performance} When fine-tuned over the entire training set, QAT usually maintains better quantized performance than PTQ. From Figure~\ref{fig:direct_ptq}, the performances of QAT are close to FP results, and remain steady across different bit-widths, i.e., W4A8, W2A8 and W2A4. However, the performances of PTQ drop significantly, which has been the main concern to address.

In summary, REM-based PTQ is superior to QAT with regard to training efficiency, memory overhead, and data accessibility. Nevertheless, it is still often less preferred than QAT due to its
severe performance drop especially for low quantization bit-width~\cite{zafrir2019q8bert,shen2020qbert,zhang2020ternarybert}.
In this paper, we aim at improving the performance of post-training quantization for PLMs, while preserving its merits of fast training, light memory overhead, and data consumption.

\section{Methodology}
\label{sec:method}

\begin{figure*}[t]
    \centering
    \includegraphics[width=0.95\textwidth]{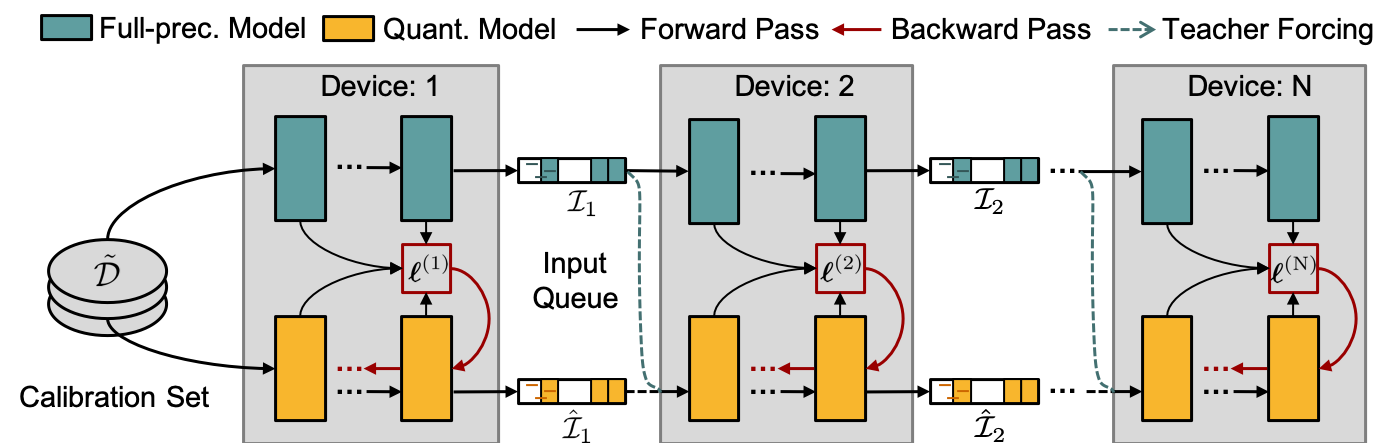}
    \caption{The overview of the proposed module-wise reconstruction error minimization~(MREM). 
    We partition both the full-precision model and quantized model into multiple modules and put these modules on different computing devices. By sampling tensors from the input queue, each module can be trained locally without waiting for its predecessors. Teacher forcing is applied to mitigate the issue of reconstruction error propagation on the quantized module.}
    \label{fig:model}
\end{figure*}

In this section, we propose our solution to improve the post-training quantization of Transformer-based PLMs. 
We first extend the existing reconstruction error minimization from the layer-wise to the module-wise granularity 
to fit Transformer models.
Secondly, based on the module partition, we further design a new parallel training strategy that further speeds up 
the PTQ pipeline. An overview of our solution can be found in Figure~\ref{fig:model}.

\subsection{Module-wise Reconstruction Error Minimization}
\label{sec:module_kd}


We propose a new PTQ solution called \textit{module-wise reconstruction error minimization}~(MREM) for PLMs.
Existing REM~\cite{nagel2020up} solves  Equation~\eqref{eq:ptq_objective} for each matrix multiplication.
However, a standard transformer layer in PLMs consists of a Multi-Head Attention (MHA) and a Feed-Forward Network (FFN),
both of which contain a number of matrix multiplications that are coupled together. 
Greedily tackling each matrix multiplication in REM can thus lead to suboptimal quantized networks.
Moreover, the insufficiently minimized reconstruction error shall propagate and enlarge along with transformer layers, 
and finally deteriorate the network output~\cite{chen2019deep,bai2020few}.



Towards that end, the proposed module-wise reconstruction error minimization admits larger granularity by 
jointly optimizing all the coupled linear layers inside each module.
Specifically, given a PLM with $L$ transformer layers, embedding layers and the classification head, 
we partition them into $N$ modules, where the $n$-th module include $[l_{n}, l_{n+1})$ transformer layers with $l_n$ 
being the first layer of this module\footnote{Note that the embedding layers and the classification head are incorporated in the first and last module respectively.}. 
MREM aims at minimizing the joint reconstruction errors between all quantized FFN output $\hat{\ms f}_l$ in the module from their full-precision counterpart $\ms f_l$ as follows:
\begin{align}
\label{eq:modulewise_kd}
\min_{\m w_n, \m s_n} \ell^{(n)} = \sum_{l\in [l_{n}, l_{n+1})} \|\hat{\ms f}_l - \ms f_l\|^2,
\end{align}
where $\m w_n$ and $\m s_n$ are all learnable parameters and quantization step sizes in the $n$-th module. 
Similar to REM, MREM can be optimized sequentially: 
given previously trained modules, only parameters and quantization step sizes in the current module are updated. 
Besides the grouped Transformer layers, we also minimize the MSE loss in the Transformer embedding and output logits respectively.

Note that the number of modules $N$ can be adjusted depending on the memory constraint of computing resources. 
When $N=1$, this reduces to intermediate-layer knowledge distillation~\cite{jiao2020tinybert}, which can be memory-demanding when quantizing large PLMs on a single GPU.

\subsection{Accelerated Parallel Training}

Based on the proposed MREM, we propose a new model parallel strategy to further accelerate the training.
As shown in Figure~\ref{fig:model}, we put different modules on individual computing devices. 
A set of \textbf{input queues} $\boldsymbol{\mathcal{I}} = \{\boldsymbol{\mathcal{I}}_1, ..., \boldsymbol{\mathcal{I}}_{N-1}\}$ is deployed between each pair of adjacent modules. For the $n$-th module, the queue collects its output of the most recent $t_0$ steps, i.e., $\boldsymbol{\mathcal{I}}^t_{n}=\{\ms f^{t}_{l_n}, \ms f^{t-1}_{l_n}, ..., \ms f^{t-t_0+1}_{l_n}\}$.
Meanwhile, the $(n+1)$-th module can always sample with replacement $\ms f_{l_{n}} \sim \boldsymbol{\mathcal{I}}_{n}^t$ from the queue without waiting for the $n$-th module. Similar rules hold for the quantized module and their input queues $\boldsymbol{\hat{\mathcal{I}}}$ as well.
The design of the input queue resembles \textit{stale synchronous parallel}~\cite{ho2013more} which stores the stale output in a local cache so as to reduce the waiting time among workers, where $t_0$ is the stale threshold.

The training workflow is as follows.
Initially, every module is computed one after another in the first $t_0$ step to fill in the input queue, after which parallel training takes place.
Then the module samples input from the queue and calculates the loss $\ell^{(n)}$ correspondingly for $n=1,...,N$. 
Meanwhile, the input queue is also updated with the rule of \textit{first-in-first-out} throughout the training.
In the backward pass, we constrain the gradients to propagate locally within each module, without affecting its predecessors.
Such a design can avoid the load imbalance issue from straggler modules, bringing nearly the theoretical $N\times$ speed-up.

\subsubsection{Annealed Teaching Forcing}
\label{sec:teach_enforce}
Since all modules proceed with training simultaneously instead of the sequential manner, the next module takes the output from the queue before its predecessor is fully optimized. Therefore, the reconstruction error from the predecessor is propagated to the following modules before it is sufficiently minimized.



Inspired by teacher forcing~\cite{williams1989learning} in training recurrent networks, the output $\ms f_{l_n}$ from the $n$-th full-precision module naturally serves as the clean input to the $(n+1)$-th quantized module to substitute $\hat{\ms f}_{l_n}$. Thus $\ms f_{l_n}$ stops the propagation of the reconstruction error accumulated on the quantized module. Nevertheless, such an approach breaks the connection to previous quantized modules and may suffer from forward inconsistency between training and inference~\cite{bai2020few} on the quantized model. 
To achieve a proper trade-off, we take the convex combination between the full-precision $\ms f_{l_n}$ and quantized $\hat{\ms f}_{l_n}$ as follows:
\begin{align}
    \label{eq:cross}
    \tilde{\ms f}_{l_{n}} = \lambda \ms f_{l_{n}} + (1-\lambda) \hat{\ms f}_{l_{n}}, \quad \lambda \in [0, 1],
\end{align}
where the hyperparameter $\lambda$ controls the strength of teacher forcing. 
$\lambda=1$ gives the full correction of reconstruction error but with forward inconsistency, while $\lambda=0$ reduces to the conventional setting that suffers from the propagated reconstruction error. We adopt a linear decay strategy for $\lambda$: $\lambda_t = \max(1 - \frac{t}{T_0},\ 0)$, where $T_0$ is the preset maximum steps of the decay.
Intuitively, a large $\lambda$ is desired at the beginning when each module is rarely optimized. Later, a small $\lambda$ is preferred to transit to normal training such that the forward inconsistency can be bridged. The remaining $T-T_0$ steps stick to normal training without teacher forcing, so as to make each quantized module adapt to its own predecessors.


\subsubsection{Comparison with Pipeline Parallelism}
\label{sec:compare_with_gpipe}
Notably, our MREM with stale synchronous parallel is different from the recent pipeline parallel~\cite{huang2018gpipe,narayanan2019pipedream}. 
Pipeline parallel adopts end-to-end training with synchronous updates between adjacent modules, which gives rise to bubble time on computing devices.
While GPipe~\cite{huang2018gpipe} divides the original data batch into $M$ pipelined micro-batches, it still has the bubble time of  $O(\frac{N-1}{N + M - 1})$ under $N$ partitions.
On the one hand, a larger $N$ or smaller $M$ would increase the bubble time. On the other hand, a larger $M$ leads to smaller batches that still cannot fully exploit the computing power, which again affects the acceleration rate.
Differently, our parallel strategy conducts local training with stale synchronous updates of the module. Hence there is negligible bubble time as long as the straggler is faster than the staleness threshold $t_0$, which can be easily satisfied with balanced module partitions or larger  $t_0$.

\begin{algorithm}[t]
\caption{Efficient Post-training Quantization.}
\label{alg:main}
\begin{algorithmic}[1]
\small
\Procedure{Main}{}:
    \vspace{0.5ex}
    \State{Partition the PLM into $N$ modules}
    \vspace{0.3ex}
    \State{Fill in the input queues $\boldsymbol{\mathcal{I}}$, $\hat{\boldsymbol{\mathcal{I}}}$}
    \vspace{0.5ex}
    \For{$n$ in $1,\ ..., N$}
        \vspace{0.35ex}
        \State{$\triangleright$\ \ \textit{run in parallel}}
        \vspace{0.35ex}
	    \While{$t<T$}
	    \vspace{0.5ex}
        \State{$\ms f_{l_{n-1}} \sim \boldsymbol{\mathcal{I}}_{n-1}^t$,\ \  $\hat{\ms f}_{l_{n-1}} \sim \hat{\boldsymbol{\mathcal{I}}}_{n-1}^t$}
        \vspace{1ex}
        \State{$\ms f_{l_n}^t, \hat{\ms f}_{l_n}^t\leftarrow $\ \textsc{MREM}\ ($\ms f_{l_{n-1}}, \hat{\ms f}_{l_{n-1}}, t$)}
        \vspace{0.5ex}
        \State{Update $\boldsymbol{\mathcal{I}}_n^t$, $\hat{\boldsymbol{\mathcal{I}}}_n^t$ with $\ms f_{l_n}^t, \hat{\ms f}_{l_n}^t$}
        \EndWhile
    \EndFor
    \vspace{0.5ex}
    \State{\textbf{return} the Quantized PLM}
\EndProcedure
\end{algorithmic}
\end{algorithm}




Finally, an overview of the proposed parallel module-wise reconstruction error minimization is shown in Algorithm~\ref{alg:main} and Algorithm~\ref{alg:sub-process}. The $\textrm{Update}(\cdot)$ in Algorithm~\ref{alg:sub-process} can be any gradient update function such as AdamW~\cite{loshchilov2018decoupled} with learning rate $\eta^t$.







\section{Experiments}
\label{sec:exp}

In this section, we empirically verify the proposed MREM for post-training quantization of PLMs. 
We first introduce the experimental setup in Section~\ref{sec:setup}. Then we present main results in Section~\ref{sec:main_results}, including comparisons with QAT and REM, as well as other existing quantization baselines.
In Section~\ref{sec:discussion}, we provide more discussions on a variety of factors in our approach, such as the effect of teacher forcing, the number of model partitions, and calibration data size.
Code will be released upon acceptance.

\begin{table*}[t]
    \centering
    \resizebox{0.98\textwidth}{!}{
    \begin{tabular}{lll|rrrll|rrrll}
    \hline\hline
    \multirow{3}{*}{} & \multirow{3}{*}{\textbf{\tabincell{c}{\#Bits\\(W-E-A)}}}  &
    \multirow{3}{*}{\textbf{\tabincell{c}{Quant\\Method}}} &
    \multicolumn{5}{c}{\textbf{BERT-base}} &
    \multicolumn{5}{c}{\textbf{BERT-large}} \\
    \cline{4-8} \cline{9-13}
    &  &  & \tabincell{c}{Time\\(min)$\downarrow$} & \tabincell{c}{Mem\\(GB)$\downarrow$} & \tabincell{c}{\# Data\\(K)$\downarrow$} & \tabincell{c}{Acc\\m(\%)$\uparrow$} & \tabincell{c}{Acc\\mm(\%)$\uparrow$} & \tabincell{c}{Time\\(min)$\downarrow$} & \tabincell{c}{Mem\\(GB)$\downarrow$} & \tabincell{c}{\# Data\\(K)$\downarrow$} & \tabincell{c}{Acc\\m(\%)$\uparrow$} & \tabincell{c}{Acc\\mm(\%)$\uparrow$} \\\hline     \multirow{13}{*}{\rot{MNLI}} & \textit{full-prec.} & N/A &  $220$ & $8.6$ & $393$ & $84.5$ & $84.9$ & $609$ & $21.5$ & $393$ & $86.7$ & $85.9$  \\
    \cline{2-13}
    & \multirow{4}{*}{4-4-8} & QAT & $1,320$ & $11.9$ & $393$ & $84.6$ & $84.9$ & $3,180$ & $29.8$ & $393$ & $86.9$ & $86.7$ \\
    &  & REM & $28$ & $2.5$ & $4$ & $73.3_{\pm 0.3}$ & $74.9_{\pm 0.2}$ & $84$ & $5.5$ & $4$ & $70.0_{\pm 0.4}$ & $71.8_{\pm 0.3}$ \\
    &  &  MREM-S & $36$ & $4.6$ & $4$ &  $83.5_{\pm 0.1}$ & $83.9_{\pm 0.1}$ &  $84$ & $10.8$ & $4$ & $86.1_{\pm 0.1}$ & $85.9_{\pm 0.1}$ \\
    &  &  MREM-P & $9$ & $3.7_{\times 4}$ & $4$ &  $83.4_{\pm 0.1}$ & $83.7_{\pm 0.1}$ & $21$ & $8.6_{\times 4}$ & $4$ & $85.5_{\pm 0.1}$ & $85.4_{\pm 0.2}$ \\
    \cline{2-13}
    & \multirow{4}{*}{2-2-8} & QAT & $882$ & $11.9$ & $393$ & $84.4$ & $84.6$ & $2,340$ & $29.8$ & $393$ & $86.5$ & $86.1$ \\
    &  & REM & $24$ & $2.5$ & $4$ & $71.6_{\pm 0.4}$ & $73.4_{\pm 0.4}$ & $64$ & $5.5$ & $4$ & $66.9_{\pm 0.4}$ & $68.6_{\pm 0.7}$ \\
    &  & MREM-S & $24$ & $4.6$ & $4$ & $82.7_{\pm 0.2}$ & $82.7_{\pm 0.2}$ & $64$ & $10.8$ & $4$ & $85.4_{\pm 0.2}$ & $85.3_{\pm 0.2} $\\
    &  & MREM-P & $6$ & $3.7_{\times 4}$ & $4$ & $82.3_{\pm 0.2}$ & $82.6_{\pm 0.2}$ & $16$ & $8.6_{\times 4}$ & $4$ & $84.6_{\pm 0.2}$ & $84.6_{\pm 0.1}$ \\
    \cline{2-13}
    & \multirow{4}{*}{2-2-4} & QAT & $875$ & $11.9$ & $393$ & $83.5$ & $84.2$ & $2,280$ & $29.8$ & $393$ & $85.8$ & $85.9$ \\
    &  & REM & $24$ & $2.5$ & $4$ & $58.3_{\pm 0.5}$ & $60.6_{\pm 0.6}$ & $64$ & $5.5$ & $4$ & $48.8_{\pm 0.6}$ & $51.4_{\pm 0.8}$ \\
    &  & MREM-S & $24$ & $4.6$ & $4$  & $81.1_{\pm 0.2}$ & $81.5_{\pm 0.2}$ & $64$ & $10.8$ & $4$ & $83.6_{\pm 0.2}$ & $83.7_{\pm 0.2}$\\
    &  & MREM-P & $6$ & $3.7_{\times 4}$ & $4$ & $80.8_{\pm 0.2}$ & $81.2_{\pm 0.2}$ & $16$ & $8.6_{\times 4}$ & $4$ & $83.0_{\pm 0.3}$ & $83.2_{\pm 0.2}$ \\
    \hline\hline
    \end{tabular}}
    \caption{Results of our proposed MREM-S and MREM-P against QAT and REM on the development set of MNLI. ``\#Bits (W-E-A)'' represents the bit-width for weights of Transformer layers, word embedding, and activations. Acc-m and Acc-mm denote accuracies on the matched and mismatched sections of MNLI respectively.}
    \label{tab:main_mnli}
\end{table*}

\begin{table*}[t]
    \centering
    \resizebox{0.98\textwidth}{!}{
    \begin{tabular}{lll|rrrll|rrrll}
    \hline\hline
    \multirow{3}{*}{} & \multirow{3}{*}{\textbf{\tabincell{c}{\#Bits\\(W-E-A)}}}  &
    \multirow{3}{*}{\textbf{\tabincell{c}{Quant\\Method}}} &
    \multicolumn{5}{c}{\textbf{BERT-base}} &
    \multicolumn{5}{c}{\textbf{BERT-large}} \\
    \cline{4-8} \cline{9-13}
    &  &  & \tabincell{c}{Time\\(min)$\downarrow$} & \tabincell{c}{Mem\\(GB)$\downarrow$} & \tabincell{c}{\# Data\\(K)$\downarrow$} & \tabincell{c}{EM (\%)$\uparrow$} & \tabincell{c}{F1 (\%)$\uparrow$} & \tabincell{c}{Time\\(min)$\downarrow$} & \tabincell{c}{Mem\\(GB)$\downarrow$} & \tabincell{c}{\# Data\\(K)$\downarrow$} & \tabincell{c}{EM (\%)$\uparrow$} & \tabincell{c}{F1 (\%)$\uparrow$} \\\hline  
    \multirow{13}{*}{\rot{SQuAD v1.1}} & \textit{full-prec.} & - & $177$ & $11.7$ & $88$ & $81.5$ & $88.7$ & $488$ & $30.4$ & $88$ & $86.9$ & $93.1$ \\
    \cline{2-13} 
    & \multirow{4}{*}{4-4-8} & QAT & $428$ & $18.4$ & $88$ & $80.2$ & $87.9$ & $\underline{1,920}$ & $\underline{27.0}$ & $88$ & $86.7$ & $93.0$ \\
    &  & REM & $65$ & $3.1$ & $4$ & $46.1_{\pm 0.5}$ & $60.0_{\pm 0.5}$ & $175$ & $7.3$ & $4$ & $68.3_{\pm 0.1}$ & $79.3_{\pm 0.1}$ \\ 
    &  & MREM-S & $76$ & $6.4$ & $4$ & $79.4_{\pm 0.1}$ & $87.2_{\pm 0.1}$ & $200$ & $14.5$ & $4$ & $86.2_{\pm 0.1}$ & $92.5_{\pm 0.1}$ \\
    &  & MREM-P & $19$ & $5.5_{\times 4}$ & $4$ &  $79.6_{\pm 0.1}$ & $87.3_{\pm 0.1}$ & $50$ & $12.3_{\times 4}$ & $4$ & $86.0_{\pm 0.1} $ & $92.4_{\pm 0.1}$  \\
    \cline{2-13}
    & \multirow{4}{*}{2-2-8} & QAT & $335$ & $18.4$ & $88$ & $79.3$ & $87.2$ & $\underline{1,200}$ & $\underline{27.0}$ & $88$ & $86.1$ & $92.5$ \\
    &  & REM & $60$ & $3.1$ & $4$ & $40.1_{\pm 0.4}$ & $55.0_{\pm 0.4}$ & $160$ & $7.3$ & $4$ & $66.4_{\pm 0.5}$ & $77.7_{\pm 0.3}$ \\
    &  & MREM-S & $60$ & $6.4$ & $4$ & $77.8_{\pm 0.2}$ & $86.0_{\pm 0.1}$ & $156$ & $14.5$ & $4$ & $85.4_{\pm 0.1}$ & $91.9_{\pm 0.1}$  \\
    &  & MREM-P & $15$ & $5.5_{\times 4}$ & $4$ & $77.7_{\pm 0.2}$ & $85.9_{\pm 0.2}$ & $39$ & $12.3_{\times 4}$ & $4$ & $85.3_{\pm 0.2}$ & $91.8_{\pm 0.1}$ \\
    \cline{2-13}
    & \multirow{4}{*}{2-2-4} & QAT & $331$ & $18.4$ & $88$ & $77.1$ & $85.9$ & $\underline{1,186}$ & $\underline{27.0}$ & $88$ & $84.7$ & $93.1$\\
    &  & REM & $60$ & $3.1$ & $4$ & $10.4_{\pm 0.2}$ & $24.6_{\pm 0.2}$ & $160$ & $7.3$ & $4$ & $28.3_{\pm 0.6}$ & $45.0_{\pm 0.5}$ \\
    &  & MREM-S & $60$ & $6.4$ & $4$ & $72.7_{\pm 0.2}$ & $82.5_{\pm 0.2}$ & $156$ & $14.5$ & $4$ & $81.4_{\pm 0.3}$ & $89.4_{\pm 0.2}$ \\
    &  & MREM-P & $15$ & $5.5_{\times 4}$ & $4$ & $73.0_{\pm 0.3}$ & $82.7_{\pm 0.2}$ & $39$ & $12.3_{\times 4}$ & $4$ & $81.8_{\pm 0.3}$ & $89.6_{\pm 0.2}$ \\
    \hline\hline
    \end{tabular}}
    \caption{Results of our proposed MREM-S and MREM-P against QAT and REM on the development set of SQuAD v1.1. ``$\underline{\quad}$'' denotes results with two gradient accumulation steps under the same total batch size due to memory constraint.}
    \label{tab:main_squad1.1}
\end{table*}

\begin{table*}[]
    \centering
    \resizebox{0.98\textwidth}{!}{
    \begin{tabular}{lll|rrrll|rrrll}
    \hline\hline
    \multirow{3}{*}{} & \multirow{3}{*}{\textbf{\tabincell{c}{\#Bits\\(W-E-A)}}}  &
    \multirow{3}{*}{\textbf{\tabincell{c}{Quant\\Method}}} &
    \multicolumn{5}{c}{\textbf{BERT-base}} &
    \multicolumn{5}{c}{\textbf{BERT-large}} \\
    \cline{4-8} \cline{9-13}
    &  &  & \tabincell{c}{Time\\(min)$\downarrow$} & \tabincell{c}{Mem\\(GB)$\downarrow$} & \tabincell{c}{\# Data\\(K)$\downarrow$} & \tabincell{c}{EM (\%)$\uparrow$} & \tabincell{c}{F1 (\%)$\uparrow$} & \tabincell{c}{Time\\(min)$\downarrow$} & \tabincell{c}{Mem\\(GB)$\downarrow$} & \tabincell{c}{\# Data\\(K)$\downarrow$} & \tabincell{c}{EM (\%)$\uparrow$} & \tabincell{c}{F1 (\%)$\uparrow$} \\\hline
    \multirow{13}{*}{\rot{SQuAD v2.0}} & \textit{full-prec.} & - & $255$ & $11.7$ & $130$ & $74.5$ & $77.7$ & $730$ & $30.4$ & $130$ & $77.7$ & $81.0$ \\
    \cline{2-13}
    & \multirow{4}{*}{4-4-8} & QAT & $662$ & $18.4$ & $130$ & $74.4$ & $77.5$ & $\underline{2,820}$ & $\underline{28.3}$ & $130$ & $77.4$ & $80.5$ \\
    &  & REM & $60$ & $3.1$ & $4$ & $53.1_{\pm 0.4}$ & $53.6_{\pm 0.4}$ & $175$ & $7.3$ & $4$ & $58.2_{\pm 0.2}$ & $61.4_{\pm 0.3}$ \\    
    &  & MREM-S & $76$ & $6.4$ & $4$ & $73.0_{\pm 0.1}$ & $76.3_{\pm 0.1}$ & $200$ & $14.5$ & $4$ & $76.4_{\pm 0.1}$ & $79.7_{\pm 0.1}$ \\
    &  & MREM-P & $19$ & $5.5_{\times 4}$ & $4$ & $72.6_{\pm 0.2}$ & $75.9_{\pm 0.2}$ & $50$ & $12.3_{\times 4}$ & $4$ & $76.3_{\pm 0.1}$ & $79.6_{\pm 0.1}$  \\
    \cline{2-13}
    & \multirow{4}{*}{2-2-8} & QAT & $508$ & $17.5$ & $130$ & $73.0$ & $76.2$ & $\underline{1,680}$ & $\underline{28.3}$ & $130$ & $76.7$ & $80.0$ \\
    &  & REM & $60$ & $3.1$ & $4$ & $51.5_{\pm 0.2}$ & $51.8_{\pm 0.2}$ & $160$ & $7.3$ & $4$ & $56.3_{\pm 0.2}$ & $59.5_{\pm 0.2}$ \\    
    &  & MREM-S & $60$ & $6.4$ & $4$ & $71.4_{\pm 0.2}$ & $74.8_{\pm 0.2}$ & $156$ & $14.5$ & $4$ & $75.4_{\pm 0.2}$ & $78.7_{\pm 0.1}$ \\
    &  & MREM-P & $15$ & $5.5_{\times 4}$ & $4$ & $70.8_{\pm 0.4}$ & $74.3_{\pm 0.4}$  &  $39$ & $12.3_{\times 4}$ & $4$ & $75.3_{\pm 0.3}$ & $78.6_{\pm 0.3}$ \\
    \cline{2-13}
    & \multirow{4}{*}{2-2-4} & QAT & $505$ & $17.5$ & $130$ & $71.4$ & $74.6$ & $\underline{1,655}$ & $\underline{28.3}$ & $130$ & $75.4$ & $78.9$ \\
    &  & REM & $60$ & $3.1$ & $4$ & $39.3_{\pm 1.5}$ & $41.4_{\pm 1.3}$ & $160$ & $7.3$ & $4$ & $42.9_{\pm 0.8}$ & $44.2_{\pm 0.7}$ \\
    &  & MREM-S & $60$ & $6.4$ & $4$ & $67.2_{\pm 0.3}$ & $70.6_{\pm 0.2}$ & $156$ & $14.5$ & $4$ & $71.3_{\pm 0.3}$ & $74.8_{\pm 0.2}$ \\
    &  & MREM-P & $15$ & $5.5_{\times 4}$ & $4$ & $66.1_{\pm 0.5}$ & $69.8_{\pm 0.5}$ & $39$ & $12.3_{\times 4}$ & $4$ & $71.5_{\pm 0.3}$ & $75.0_{\pm 0.3}$ \\
    \hline\hline
    \end{tabular}}
    \caption{Results of our proposed MREM-S and MREM-P against QAT and REM on the development set of SQuAD v2.0. ``$\underline{\quad}$'' denotes results with two gradient accumulation steps under the same total batch size due to memory constraint.}
    \label{tab:main_squad2.0}
\end{table*}

\begin{table*}[t]
	\centering
	\resizebox{0.98\textwidth}{!}{
		\begin{tabular}{lccc|llllllll|l}
    		\hline\hline
			\textbf{\tabincell{c}{Quant\\Method}} &
			\textbf{\tabincell{c}{\#Bits\\(W-E-A)}} &  \textbf{\tabincell{c}{Size\\(MB)}} & 
			\textbf{\tabincell{c}{PTQ}} &
			\textbf{\tabincell{c}{MNLI-m}} & \textbf{QQP} & \textbf{QNLI} & \textbf{SST-2} & \textbf{CoLA} & \textbf{STS-B} & \textbf{MRPC} & \textbf{RTE} &
			\textbf{Avg.}\\
			\hline
			- & \textit{full-prec.} & $418$ & - & $84.9$ & $91.4$ & $92.1$ & $93.2$ & $59.7$ & $90.1$ & $86.3$ & $72.2$ & $83.9$ \\\hline
			Q-BERT & 2-8-8 & $43$ & \xmark & $76.6$ & - & - & $84.6$ & - & - & - & - & - \\			
			Q-BERT & 2/4-8-8 & $53$ & \xmark & $83.5$ & - & - & $92.6$ & - & - & - & - & - \\
            Quant-Noise & PQ & $38$ & \xmark & $83.6$ & - & - & - & - & - & - & - & - \\
			TernaryBERT & 2-2-8 & $28$ & \xmark & $83.3$ & $90.1$ & $91.1$ & $92.8$ & $55.7$ & $87.9$ & $87.5$ & $72.9$ & $82.7$ \\
            GOBO & 3-4-32 & $43$ & \cmark & $83.7$ & - & - & - & - & $88.3$ & - & - & - \\
            GOBO & 2-2-32 & $28$ & \cmark & $71.0$ & - & - & - & - & $82.7$ & - & - & - \\
			\hline
            MREM-S & 4-4-8 & $50$ & \cmark & $83.5_{\pm 0.1}$ & $90.2_{\pm 0.1}$ & $91.2_{\pm 0.1}$ & $91.4_{\pm 0.4}$ & $55.1_{\pm 0.8}$ & $89.1_{\pm 0.1}$ & $84.8_{\pm 0.0}$ & $71.8_{\pm 0.0}$ & $82.4_{\pm 0.1}$ \\
            & 2-2-8 & $28$ & \cmark & $82.7_{\pm 0.2}$ & $89.6_{\pm 0.1}$ & $90.3_{\pm 0.2}$ & $91.2_{\pm 0.4}$ & $52.3_{\pm 1.0}$ & $88.7_{\pm 0.1}$ & $86.0_{\pm 0.0}$ & $71.1_{\pm 0.0}$ & $81.5_{\pm 0.2}$ \\
		    MREM-P & 4-4-8 & $50$ & \cmark & $83.4_{\pm 0.1}$ & $90.2_{\pm 0.1}$ & $91.0_{\pm 0.2}$ & $91.5_{\pm 0.4}$ & $54.7_{\pm 0.9}$ & $89.1_{\pm 0.1}$ & $86.3_{\pm 0.0}$ & $71.1_{\pm 0.0}$ & $82.2_{\pm 0.1}$ \\
		    & 2-2-8 & $28$ & \cmark & $82.3_{\pm 0.2}$ & $89.4_{\pm 0.1}$ & $90.3_{\pm 0.2}$ & $91.3_{\pm 0.4}$ & $52.9_{\pm 1.2}$ & $88.3_{\pm 0.2}$ & $85.8_{\pm 0.0}$ & $72.9_{\pm 0.0}$ & $81.6_{\pm 0.2}$ \\
			\hline\hline
		\end{tabular}}
	\caption{Results on the GLUE development set.
	``Size" refers to model storage in ``MB". 
	``PTQ'' indicates whether the method 
	belongs to post-training quantization. 
	``Avg." denotes the average results of all tasks.
	}
	\label{tab:compare_sota}
\end{table*}

\begin{algorithm}[t]
\caption{Module-wise Reconstruction Error Min.}
\label{alg:sub-process}
\begin{algorithmic}[1]
\small
\Procedure{MREM\ }{$\ms f_{l_{n-1}}, \hat{\ms f}_{l_{n-1}}, t$}:
\vspace{0.3ex}
    \If{$t < T_0$}
        \State{$\lambda_t \leftarrow \max(1 - \frac{t}{T_0},\ 0)$}
        \State{Compute $\tilde{\ms f}_{l_{n-1}}$ by Equation~\eqref{eq:cross}}
    \EndIf
    \State{Compute the full-precision module output $\ms f_{l_{n}}^t$}
    \State{Compute the quantized module output  $\hat{\ms f}_{l_{n}}^t$}
    \State{Compute the loss $\ell^{(n)}$  by Equation~\eqref{eq:modulewise_kd}}
    \vspace{0.4ex}
    \State{$\m w_n^{t+1} \leftarrow \textrm{Update}(\m w_n^{t}, \frac{\partial \ell^{(n)}}{\partial \m w_n^t}, \eta^t)$}
    \State{$\m s_n^{t+1} \leftarrow \textrm{Update}(\m s_n^{t}, \frac{\partial \ell^{(n)}}{\partial \m s_n^t}, \eta^t)$}
    \State{\textbf{return} $\ms f_{l_{n}}^t, \hat{\ms f}_{l_{n}}^t$}
\EndProcedure
\end{algorithmic}
\end{algorithm}

\subsection{Experimental Setup}
\label{sec:setup}

\subsubsection{Datasets and Metrics}
We evaluate post-training quantization w.r.t. both text classification on the GLUE dataset~\cite{wang2018glue}, 
and reading comprehension on SQuAD benchmarks~\cite{rajpurkar2016squad}. 
The size of calibration data is by default $|\tilde{\mathcal{D}}|=4,096$, 
with instances randomly sampled from the full training set. 
As both RTE and MRPC tasks in the GLUE benchmark contain fewer than 4,096 samples, we use their full training set on these two tasks. 
We leave the study of data size in Section~\ref{sec:discussion}. 
Each experiment is repeated ten times with different calibration sets, and both the mean and standard deviations are reported.



We use the same evaluation metrics in \cite{devlin2019bert,zhang2020ternarybert} for the development set of GLUE and SQuAD benchmarks.
For results in Section~\ref{sec:main_results}, we report  accuracies on both the matched section and mis-matched sections of MNLI, and EM~(exact match) and F1 score for SQuAD.
Additionally, we also report the training time (min), memory overhead (GB) as well as the size of the training set (K).
We also provide comparisons with other existing methods in Section~\ref{sec:compare_to_sota}, where we adopt Matthews correlation for CoLA, Spearman correlation for STS-B, and accuracy for the rest ones (i.e., RTE, MRPC, SST-2, QQP, MNLI). We also report the averaged performance on GLUE as an overview.




\subsubsection{Implementation}
We use the standardly fine-tuned BERT-base and BERT-large models\footnote{We follow the default fine-tuning hyperparameter settings in Huggingface: \textrm{https://github.com/huggingface/transformers}.} 
on downstream tasks for post-training quantization. 
We implement MREM in both the sequential training (abbv. MREM-S) in Section~\ref{sec:module_kd} and parallel training with teaching forcing (abbv. MREM-P) in Section~\ref{sec:teach_enforce}.
For each module, we train for $2,000$ steps with an initial learning rate of $1e^{-4}$ on GLUE tasks, 
and $4,000$ steps with an initial learning rate of $5e^{-5}$ on SQuAD datasets. The learning rate decays linearly as done in~\cite{jiao2020tinybert,zhang2020ternarybert}.
By default, we partition the model into 4 modules on 4 NVIDIA-V100 GPUs. The analysis of the training steps and partition numbers will be provided in Section~\ref{sec:discussion}.


For baselines, we mainly compare with QAT and REM, where the former measures how much PTQ can get close to QAT, and the latter studies the effect of objective granularity in PTQ training.
We conduct QAT following the state-of-the-art training pipeline~\cite{zhang2020ternarybert}, i.e., intermediate-layer distillation followed by prediction-layer distillation, which takes 6 training epochs in total. Detailed hyperparameter settings can be found in~\cite{zhang2020ternarybert}. 
In terms of REM, we follow the practice in \cite{nagel2020up,hubara2020improving} to minimize the reconstruction error after each matrix multiplication, as introduced in Section~\ref{sec:background}. 
For a fair comparison of each method, we use the same quantization scheme, i.e., TWN~\cite{li2016ternary} or LAQ~\cite{hou2018loss} for 2-bit and 4-bit weight quantization, and LSQ~\cite{esser2019learned} for all activation quantization. Unlike QAT that picks the best model based on the development set results, MREM is only tested once after training, which ensures data security of the development set. 
We leave the comparison with more existing quantization approaches in Section~\ref{sec:compare_to_sota}.

\subsection{Main Results: Comparison with QAT and REM}
\label{sec:main_results}

We first compare MREM-S and MREM-P with QAT and REM over MNLI and SQuAD benchmarks. 
We take BERT-base and BERT-large as backbone PLMs for quantization. 
The results on MNLI, SQuADv1.1 and SQuADv2.0  
are summarized in Table~\ref{tab:main_mnli}, Table~\ref{tab:main_squad1.1} and Table~\ref{tab:main_squad2.0} respectively.
We summarize the results from the four dimensions mentioned in Section~\ref{sec:why_ptq}.

\subsubsection{Performance}
It can be found that our proposed MREM-S improves the performance of REM significantly 
given the same training time, and is much closer to QAT.
For instance, according to in Table~\ref{tab:main_mnli}, MREM-S with 4-bit weight quantization on BERT-base and BERT-large achieves accuracies of $83.5\%_{\pm 0.1}$ and $86.1\%_{\pm 0.1}$ on the matched section of MNLI, which is on average ${10.2}\%\uparrow$ and ${16.1}\%\uparrow$ better than REM, and only ${1.1}\%\downarrow$ and ${0.8}\%\downarrow$ inferior to QAT, respectively. 
With REM, BERT-base sometimes even outperforms BERT-large on MNLI. We speculate that this is due to the suboptimal solutions in REM that lead to propagated reconstruction error when more neurons or transformer layers are stacked in BERT.

Moreover, with all modules trained in parallel, MREM-P 
is still close to or only slightly inferior to MREM-S. 
From  results of SQuAD v1.1 in Table~\ref{tab:main_squad1.1}, MREM-P can even outperform MREM-S with the ``W2-E2-A4'' quantized BERT-large model (i.e., the EM score and F1 score are on average ${0.4}\%\uparrow$ and ${0.2}\%\uparrow$ respectively).



\subsubsection{Training Time}
Our proposed MREM also enjoys significantly less training time than QAT. 
For instance, MREM only takes $84$ minutes for $4$-bit weight quantized training on the BERT-large over MNLI, which is about ${38}\times$ faster than QAT and ${7}\times$ faster than full-precision fine-tuning. 
When compared with REM, MREM does not need to cache the output after every matrix multiplication, 
which admits more iterations given the same amount of time. We shall discuss this further in Section~\ref{sec:longer_rem}.
Moreover, when armed with the proposed parallel training, MREM-P is further $\textbf{4}\times$ faster than MREM-S, which achieves the theoretical linear speedup on $4$ GPUs. These together bring more than $\textbf{150}\times$ reduction of training time when compared with QAT.

\begin{table*}[t]
	\centering
	\makeatletter\def\@captype{table}\makeatother
	\begin{minipage}{0.495\textwidth}
	\resizebox{0.95\textwidth}{!}{
	\begin{tabular}{cc|cc|cc}
		\hline\hline
		    \multirow{2}{*}{\textbf{\tabincell{c}{\#Bits\\(W-E-A)}}} &
		    \multirow{2}{*}{\textbf{\# Steps}} & \multicolumn{2}{c}{\textbf{BERT-base}} & \multicolumn{2}{c}{\textbf{BERT-large}} \\
		    \cline{3-4}\cline{5-6}
		    &  & \textbf{w/o TF} & \textbf{w/ TF} &  \textbf{w/o TF} & \textbf{w/ TF} \\\hline
		    \multirow{4}{*}{2-2-8} & $250$ & $79.6_{\pm 0.3}$ & $80.7_{\pm 0.2}$ & $82.1_{\pm 0.4}$ & $83.1_{\pm 0.2}$ \\
             & $500$ &  $81.0_{\pm 0.3}$ & $81.6_{\pm 0.2}$ & $83.4_{\pm 0.3}$ & $84.1_{\pm 0.3}$ \\
             & $2,000$ & $82.2_{\pm 0.2}$ & $82.7_{\pm 0.2}$ & $84.3_{\pm 0.3}$ & $84.6_{\pm 0.2}$ \\
             & $4,000$ & $82.3_{\pm 0.3}$ & $82.5_{\pm 0.2}$ & $84.5_{\pm 0.2}$ & $84.7_{\pm 0.2}$ \\
             \hline
		    \multirow{4}{*}{2-2-4} & $250$ & $73.9_{\pm 0.5}$ & $77.3_{\pm 0.4}$ & $76.5_{\pm 0.9}$ & $79.3_{\pm 0.4}$ \\
             & $500$ &  $77.9_{\pm 0.2}$ & $79.0_{\pm 0.2}$ & $80.0_{\pm 0.5}$ & $81.4_{\pm 0.2}$ \\
             & $2,000$ & $80.4_{\pm 0.2}$ & $80.8_{\pm 0.2}$ & $82.5_{\pm 0.4}$ & $83.0_{\pm 0.3}$ \\
             & $4,000$ & $80.7_{\pm 0.2}$ & $81.0_{\pm 0.2}$ & $83.1_{\pm 0.1}$ & $83.3_{\pm 0.3}$ \\
             \hline\hline
		\end{tabular}}
    \captionsetup{margin=0.15cm}
    \vspace{-0.2ex}
    \caption{Ablation studies of teacher forcing at different training steps over MNLI-m.}
    \label{tab:teacher_forcing}
	\end{minipage}
	\hfill
	\begin{minipage}{0.495\textwidth}
	\resizebox{0.99\textwidth}{!}{
	\begin{tabular}{cl|rrrcc}
	\hline\hline
	    \textbf{\tabincell{c}{\#Bits\\(W-E-A)}} &
	    \textbf{\tabincell{c}{Quant\\Method}} &
	    \textbf{\tabincell{c}{\#\ Steps}} &
	    \textbf{\tabincell{c}{Time\\(min)$\downarrow$}} & \textbf{\tabincell{c}{Mem\\(G)$\downarrow$}} & \textbf{\tabincell{c}{Acc\\m(\%)$\uparrow$}} & \textbf{\tabincell{c}{Acc\\mm(\%)$\uparrow$}} \\\hline
	    \multirow{3}{*}{4-4-8} & REM & $200$ & $36$ & $2.5$ & $73.3_{\pm 0.3}$ & $74.9_{\pm 0.2}$ \\
         & REM & $2,000$ & $319$ & $2.5$ & $81.8_{\pm 0.2}$ & $82.5_{\pm 0.1}$ \\
         & MREM-S & $2,000$ & $36$ & $4.6$ &  $83.5_{\pm 0.1}$ & $83.9_{\pm 0.1}$ \\\hline 
	    \multirow{3}{*}{2-2-8} & REM & $200$ & $24$ & $2.5$ & $71.6_{\pm 0.4}$ & $73.4_{\pm 0.4}$ \\
         & REM & $2,000$ & $213$ & $2.5$ & $78.7_{\pm 0.2}$ & $79.2_{\pm 0.2}$ \\
         & MREM-S & $2,000$ & $24$ & $4.6$ & $82.7_{\pm 0.2}$ & $82.7_{\pm 0.2}$ \\\hline
	    \multirow{3}{*}{2-2-4} & REM & $200$ &  $24$ & $2.5$ & $58.3_{\pm 0.5}$ & $60.6_{\pm 0.6}$ \\
         & REM & $2,000$ & $213$ & $2.5$ & $73.0_{\pm 0.3}$ & $74.4_{\pm 0.4}$ \\
         & MREM-S & $2,000$ & $24$ & $4.6$ & $81.1_{\pm 0.2}$ & $81.5_{\pm 0.2}$ \\
         \hline\hline
	\end{tabular}}
	\captionsetup{margin=0.15cm}
	\vspace{-0.5ex}
	\caption{Comparison of REM with our MREM on BERT-base over MNLI.}
	\label{tab:REM_MREM}
	\end{minipage}
\end{table*}

\begin{figure*}[t]
	\subfigure[Module-1 (250 Steps).]{
	    \includegraphics[width=0.23\textwidth]{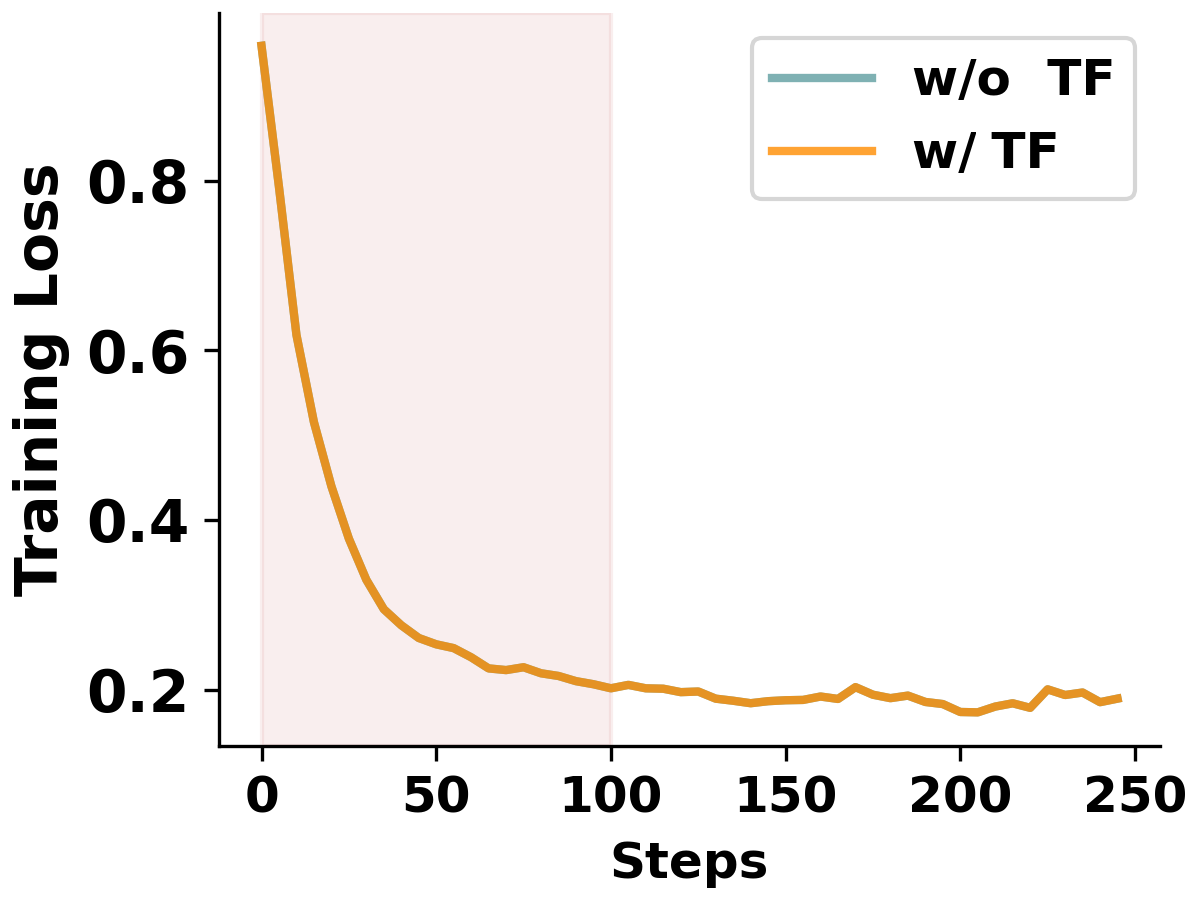}
	    \label{fig:250_p1}
	}
	\subfigure[Module-2 (250 Steps).]{
	    \includegraphics[width=0.23\textwidth]{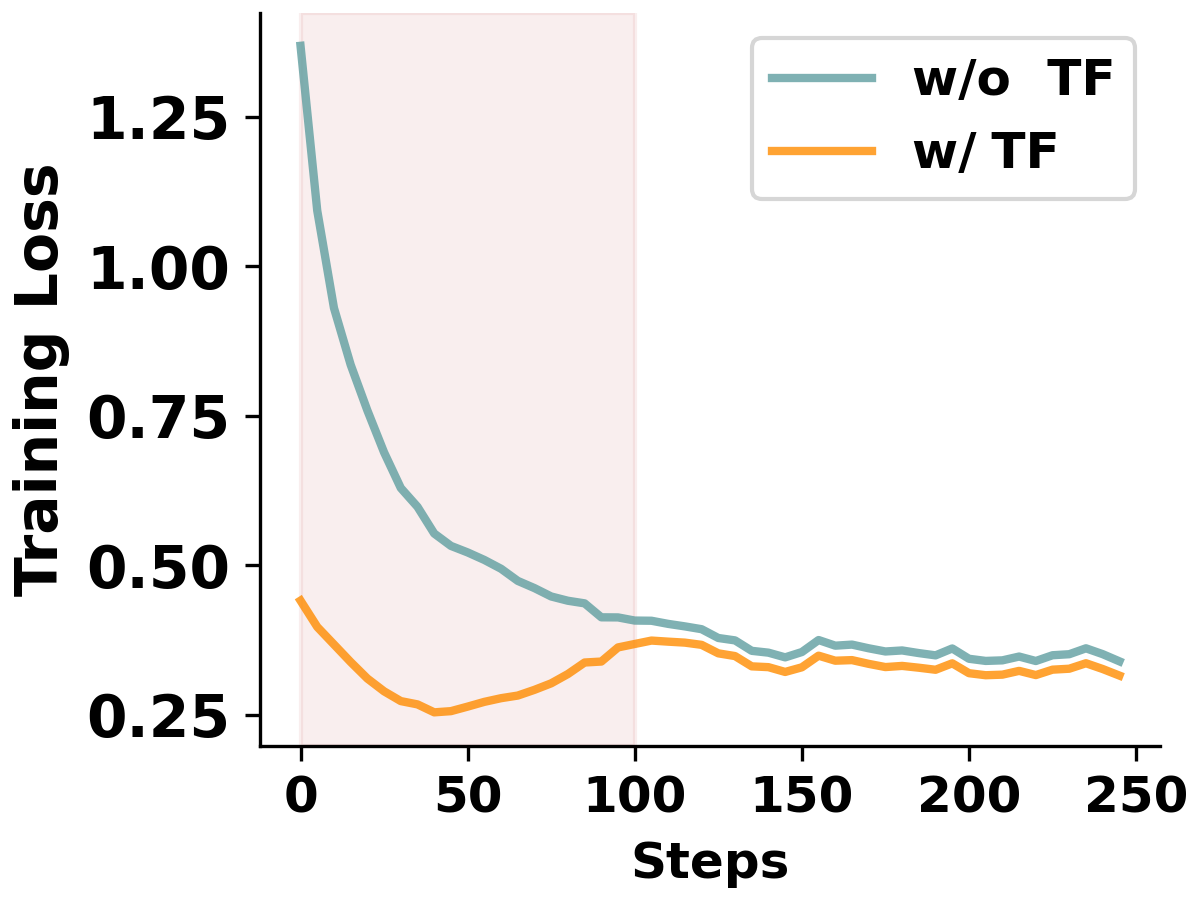}
	    \label{fig:250_p2}
	}
	\subfigure[Module-3 (250 Steps).]{
	    \includegraphics[width=0.23\textwidth]{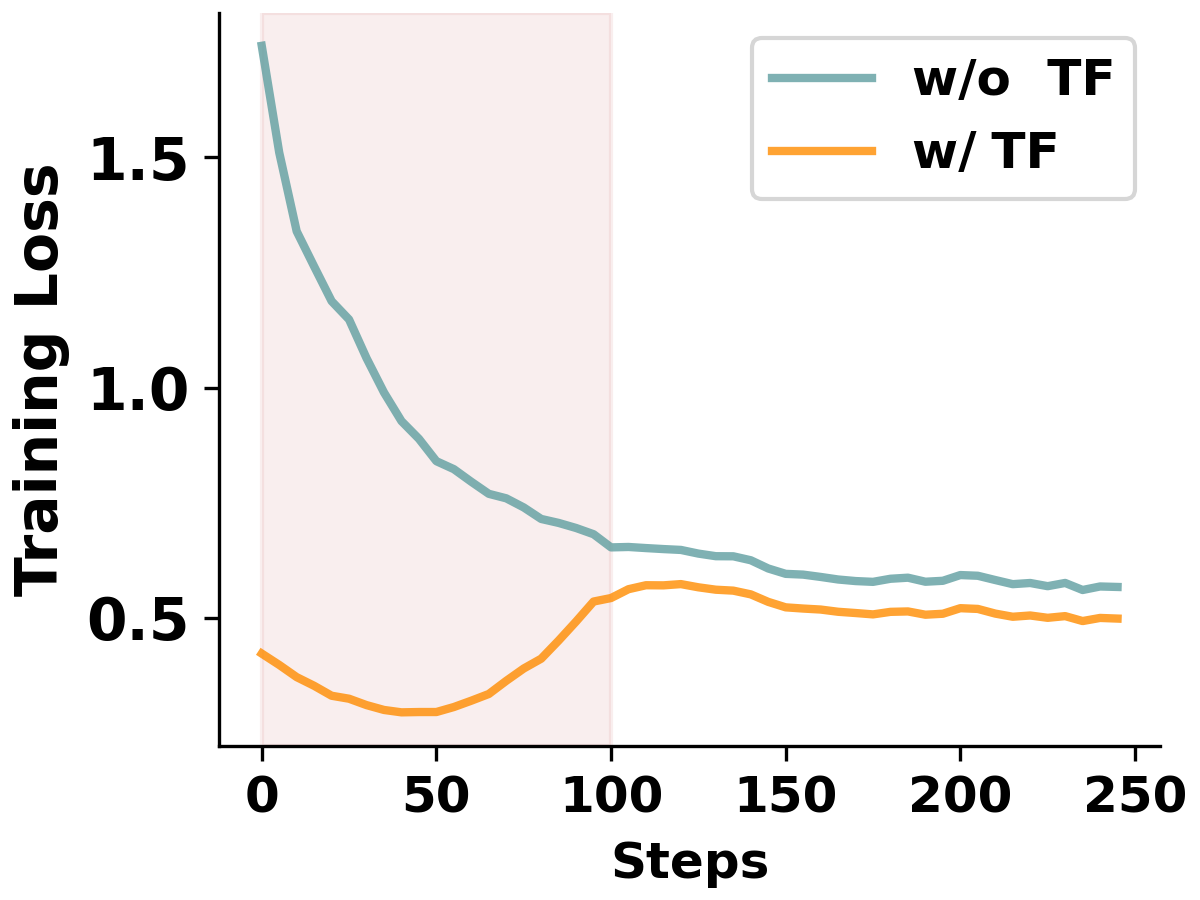}
	    \label{fig:250_p3}
	}
	\subfigure[Module-4 (250 Steps).]{
	    \includegraphics[width=0.23\textwidth]{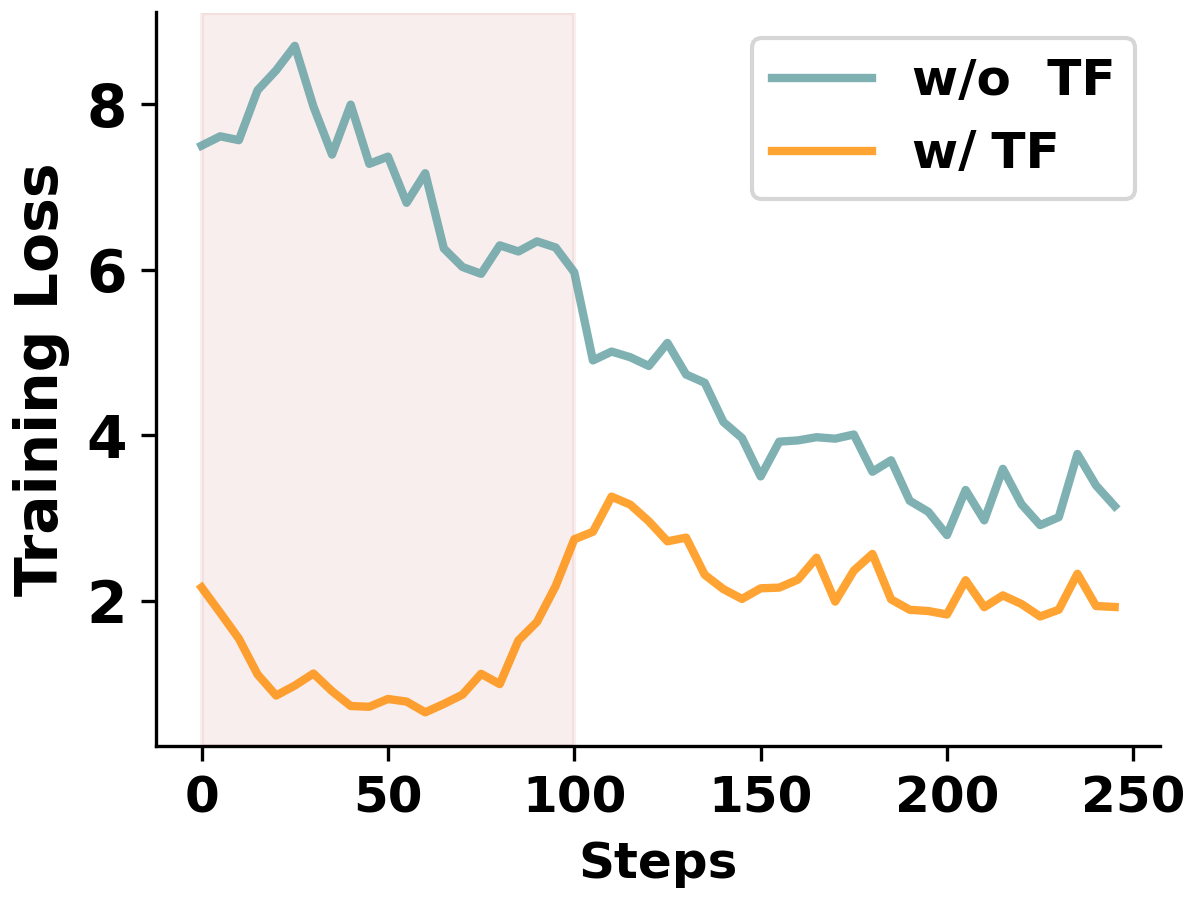}
	    \label{fig:250_p4}
	}
	\subfigure[Module-1 (2,000 Steps).]{
	    \includegraphics[width=0.23\textwidth]{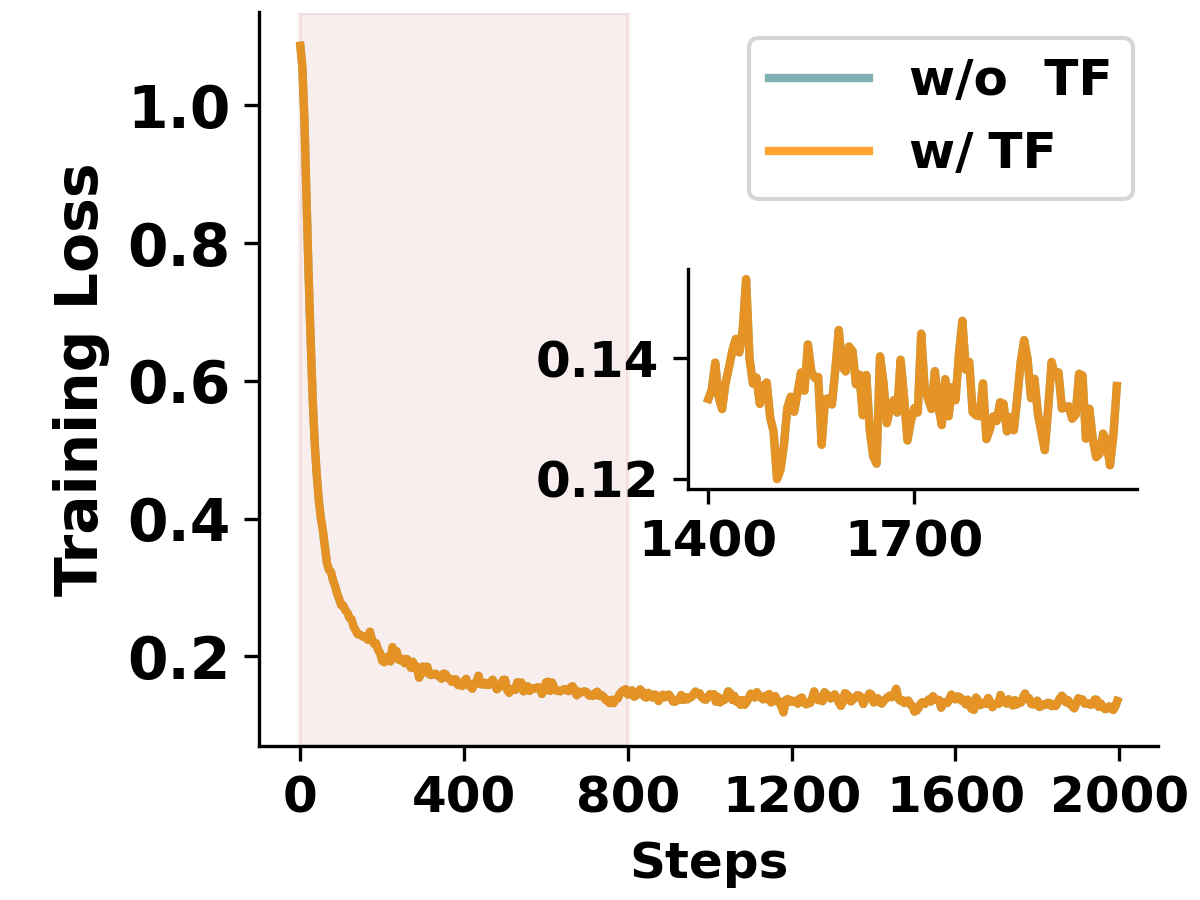}
	    \label{fig:2000_p1}
	}
	\subfigure[Module-2 (2,000 Steps).]{
	    \includegraphics[width=0.23\textwidth]{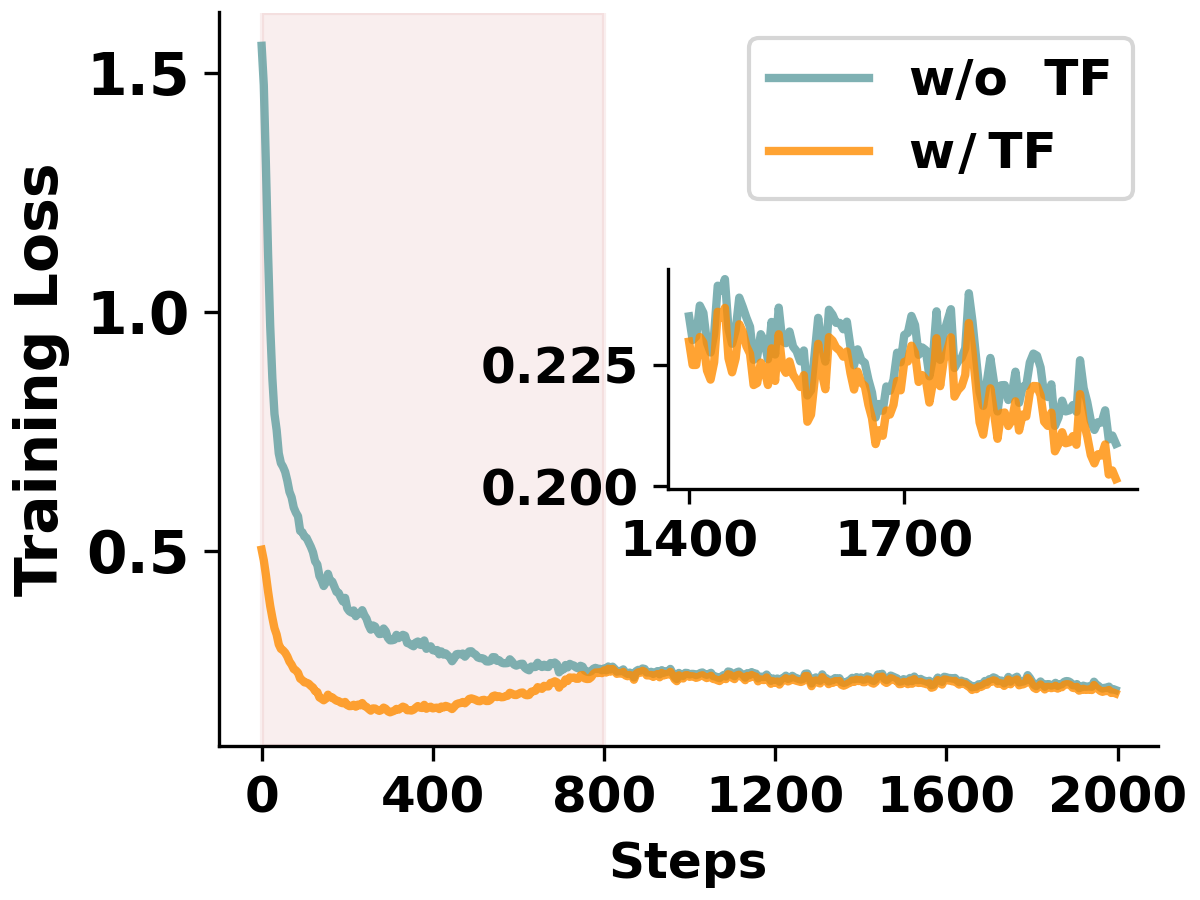}
	    \label{fig:2000_p2}
	}
	\subfigure[Module-3 (2,000 Steps).]{
	    \includegraphics[width=0.23\textwidth]{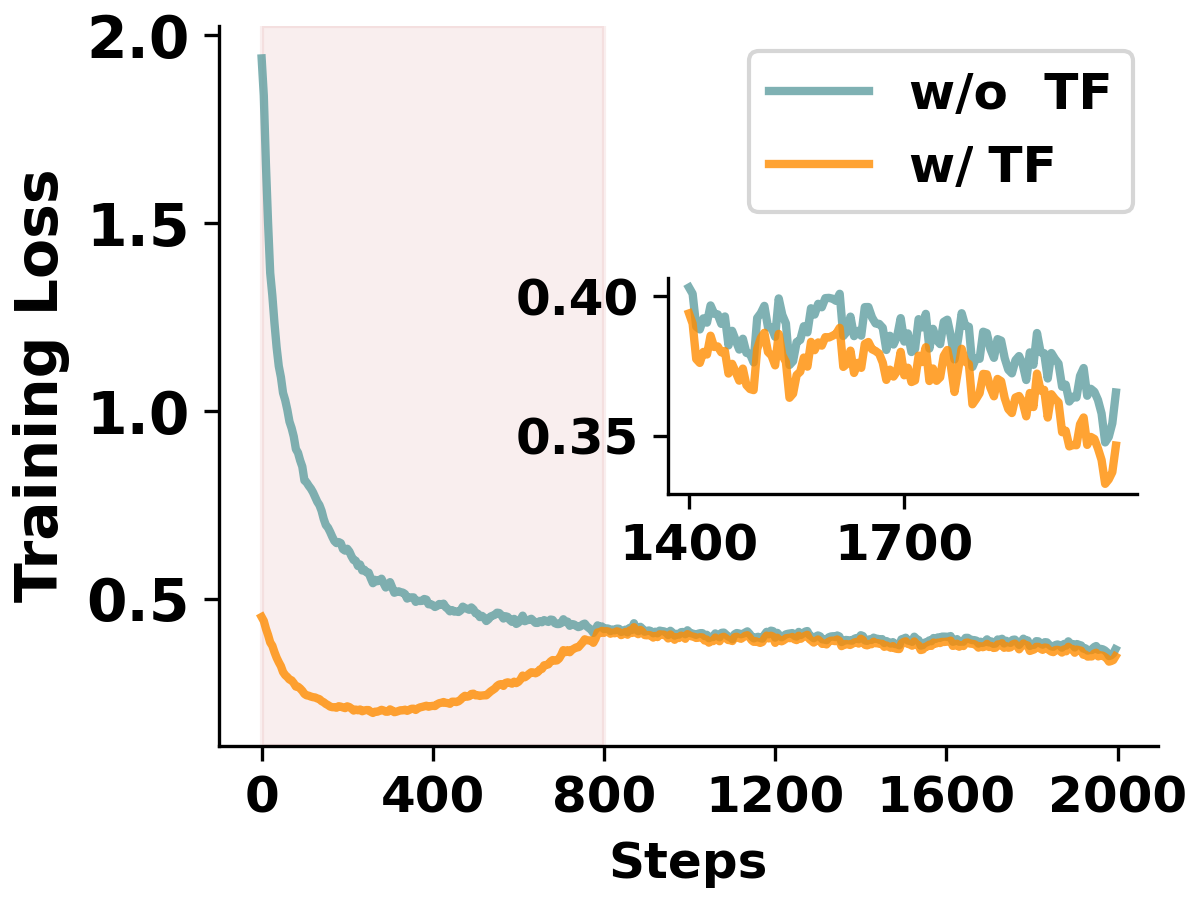}
	    \label{fig:2000_p3}
	}
	\subfigure[Module-4 (2,000 Steps).]{
	    \includegraphics[width=0.23\textwidth]{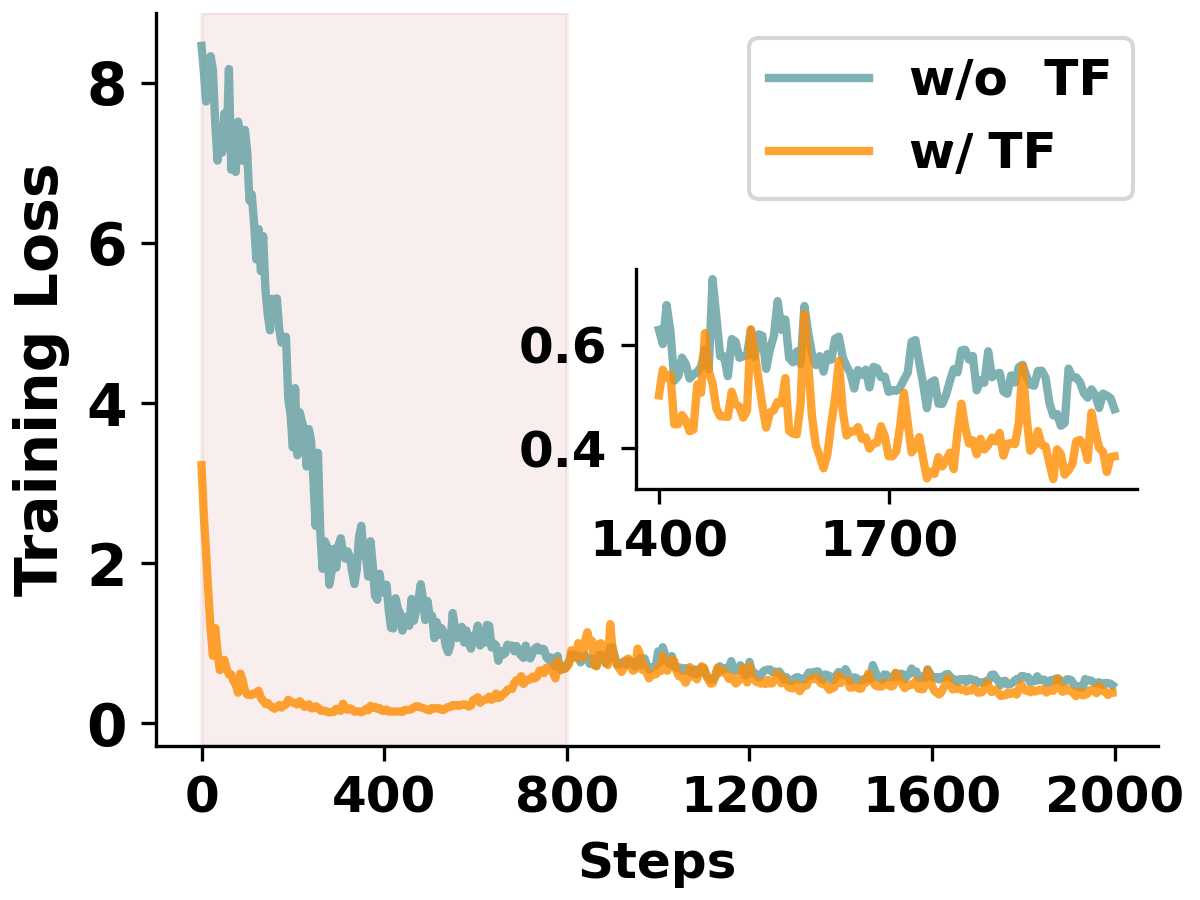}
	    \label{fig:2000_p4}
	}	
    \vspace{-1ex}
	\caption{The training loss curves with and without teacher forcing (TF) in MREM-P. The red area denotes teacher forcing in the first $40\%$ training steps. (a), (b), (c) and (d) in the first row are the four modules trained for 250 steps, and (e), (f), (g) and (h) in the second row are trained for 2,000 steps.}
    \label{fig:parallel_loss_curve}
\end{figure*}

\subsubsection{Memory Overhead}
While the module-wise training inevitably consumes more memory than REM, it still takes only around a third of the memory by QAT, and a half of that by the full-precision fine-tuning. 
For instance, while QAT takes $29.8$ GB memory on BERT-large, MREM only consumes $10.8$ GB memory, 
which can be even fed into a single  NVIDIA GTX 1080 Ti.
Moreover, for input with a longer sequence length (i.e., 384 tokens on the SQuAD dataset), QAT over BERT-large may suffer from memory overflow even on an NVIDIA V100 GPU with 32GB memory. QAT with gradient accumulation inevitably doubles the training time under the same total batch size (i.e., underlined figures (``\underline{\quad}'') in Table~\ref{tab:main_squad1.1} and Table~\ref{tab:main_squad2.0}). 
On the other hand, such issues can be easily mitigated in both REM and our proposed MREM.
Meanwhile, increasing the number of modules can  further decrease the memory overhead of each module, but may harm the performance, as will be discussed in Section~\ref{sec:num_module}.


\subsubsection{Data Accessibility}
Both REM and our proposed MREM follow the common practice of PTQ, relying on only $4,096$ randomly sampled training instances on both MNLI and SQuAD, which is a tiny fraction of the original dataset used in QAT. We shall provide more discussion on the effect of calibration size in Section~\ref{sec:discussion}.


\textbf{In summary}, our MREM-S improves post-training quantization on PLMs significantly, while still enjoys fast training, light memory overhead, and data security. Moreover, with parallel training, the proposed MREM-P further strengthens the advantages of PTQ without an apparent performance drop.

\subsection{Main Results: Comparison with Existing Methods}
\vspace{-1ex}
\label{sec:compare_to_sota}
In the next, we compare our MREM with a number of existing state-of-the-art BERT quantization methods. They include various QAT approaches such as Q-BERT~\cite{shen2020qbert},  Quant-Noise~\cite{fan2020training} and TernaryBERT~\cite{zhang2020ternarybert}, as well as the PTQ baseline GOBO~\cite{zadeh2020gobo}.
Their results are taken from the original papers, respectively. 

From Table~\ref{tab:compare_sota}, both our proposed MREM-S and MREM-P outperform existing PTQ approaches in most cases, and even achieve results close to QAT approaches.
For example, the ``W4-E4-A8'' quantized MREM-S and MREM-P have the averaged accuracies of $83.5\%$ and $83.4\%$ on MNLI respectively, both of which are on par with ``W2/4-E8-A8'' quantized Q-BERT.
In terms of the ``W2-E2-A8'' quantized models, our MREM-S and MREM-P surpass GOBO by $\textbf{11.7}\%\uparrow$ and $\textbf{11.3}\%\uparrow$ on MNLI-m respectively.

\subsection{Discussions}
\label{sec:discussion}

In this section, we provide further discussions to better understand the proposed approach. By default, all experiments in this section are based on the BERT-base model over the MNLI dataset.


\begin{figure*}[t]
	\subfigure[Number of Modules and Memory Overhead.]{
	    \includegraphics[width=0.235\textwidth]{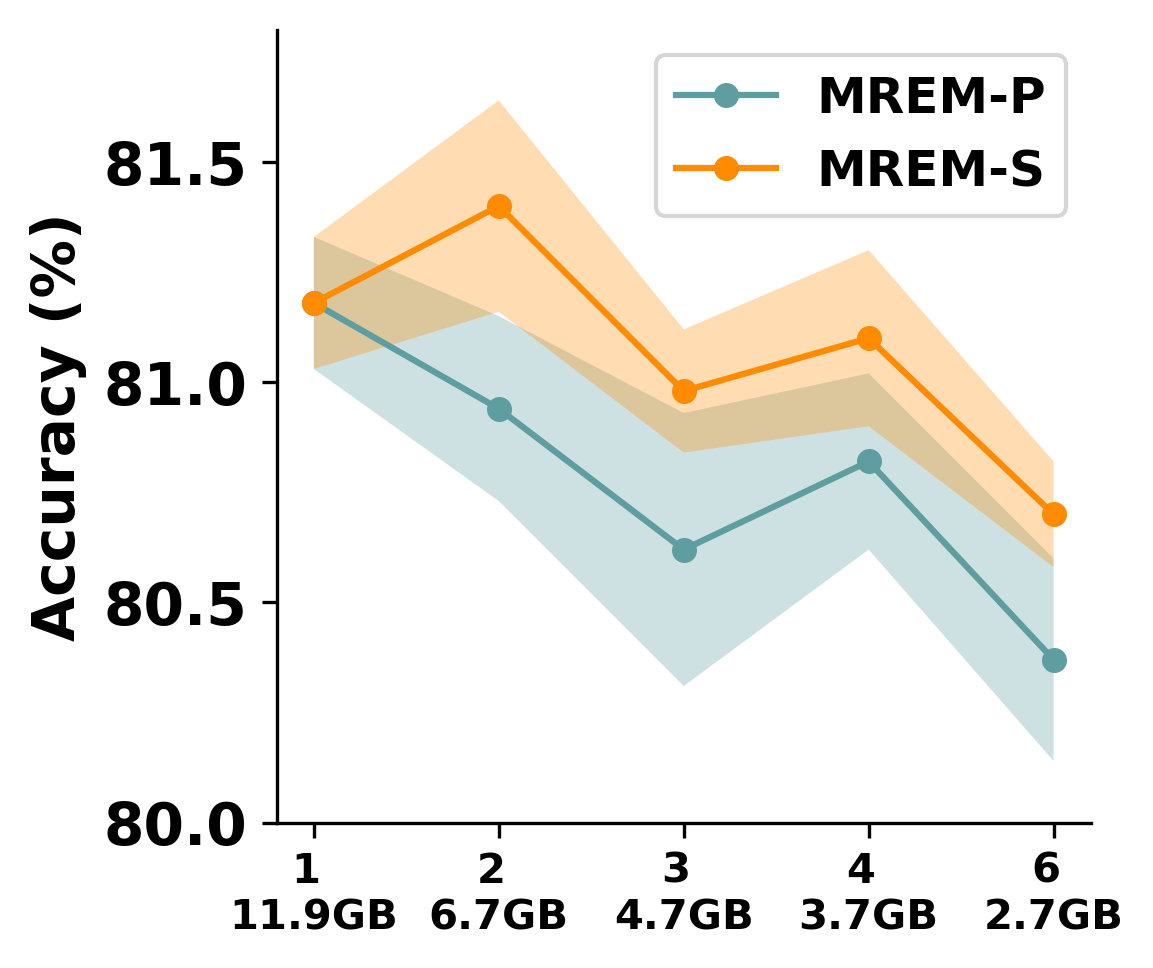}
	    \label{fig:num_partition}
	}
	\subfigure[Size of Calibration Data.]{
	    \includegraphics[width=0.215\textwidth]{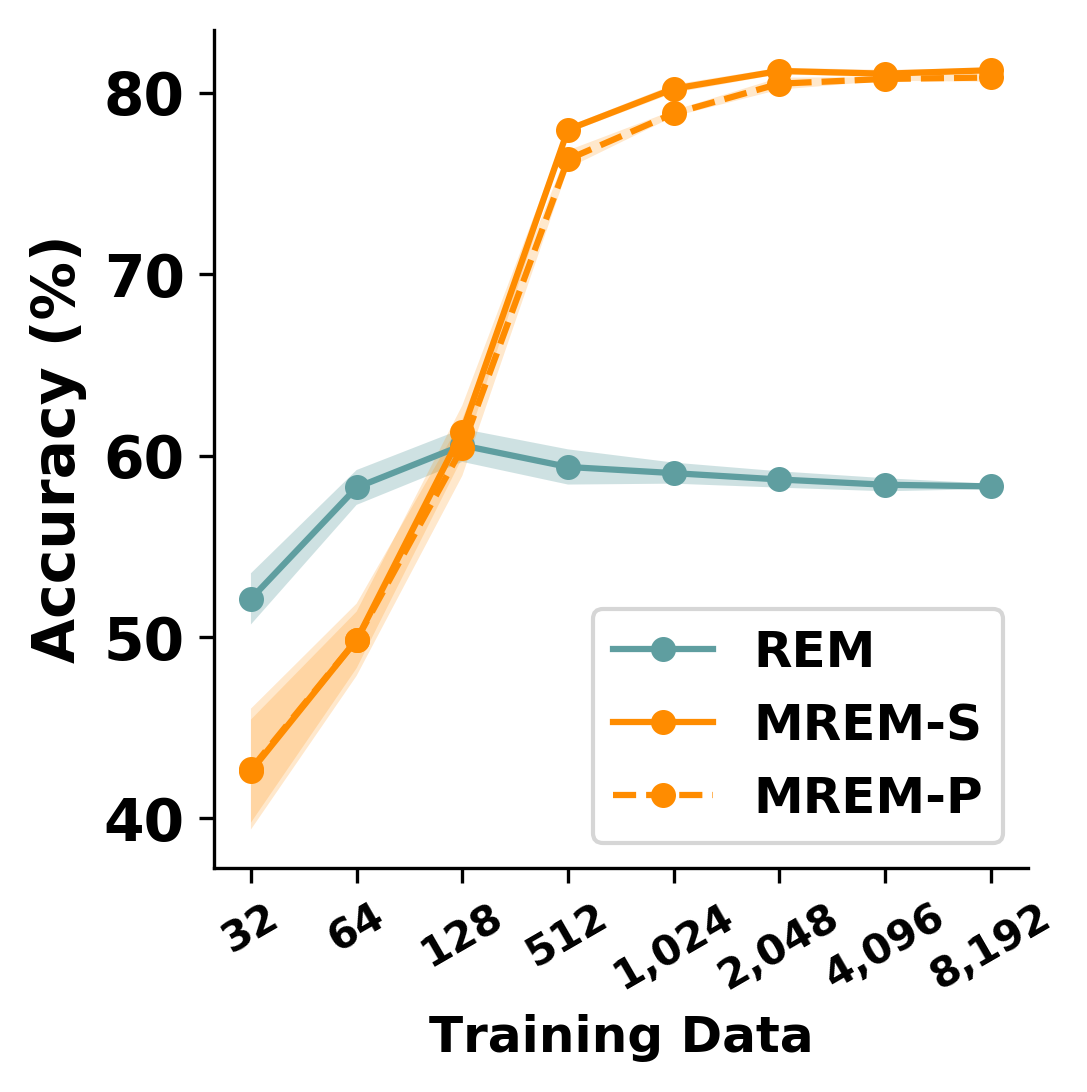}
	    \label{fig:data_size}
	}		
	\subfigure[Error Propagation~(A8).]{
	    \includegraphics[width=0.22\textwidth]{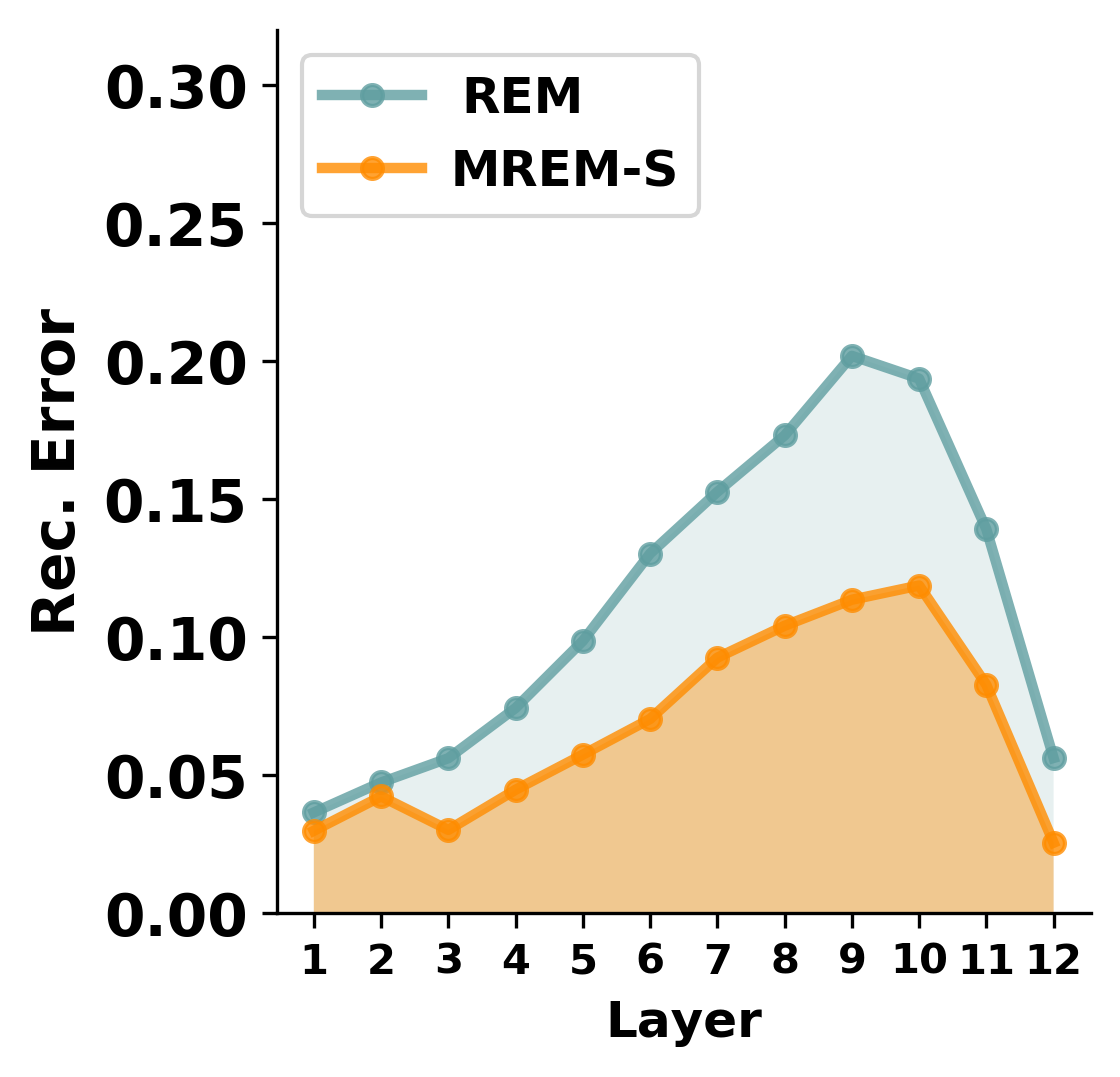}
	    \label{fig:error_propagate_a8}
	}	
	\subfigure[Error Propagation (A4).]{
	    \includegraphics[width=0.22\textwidth]{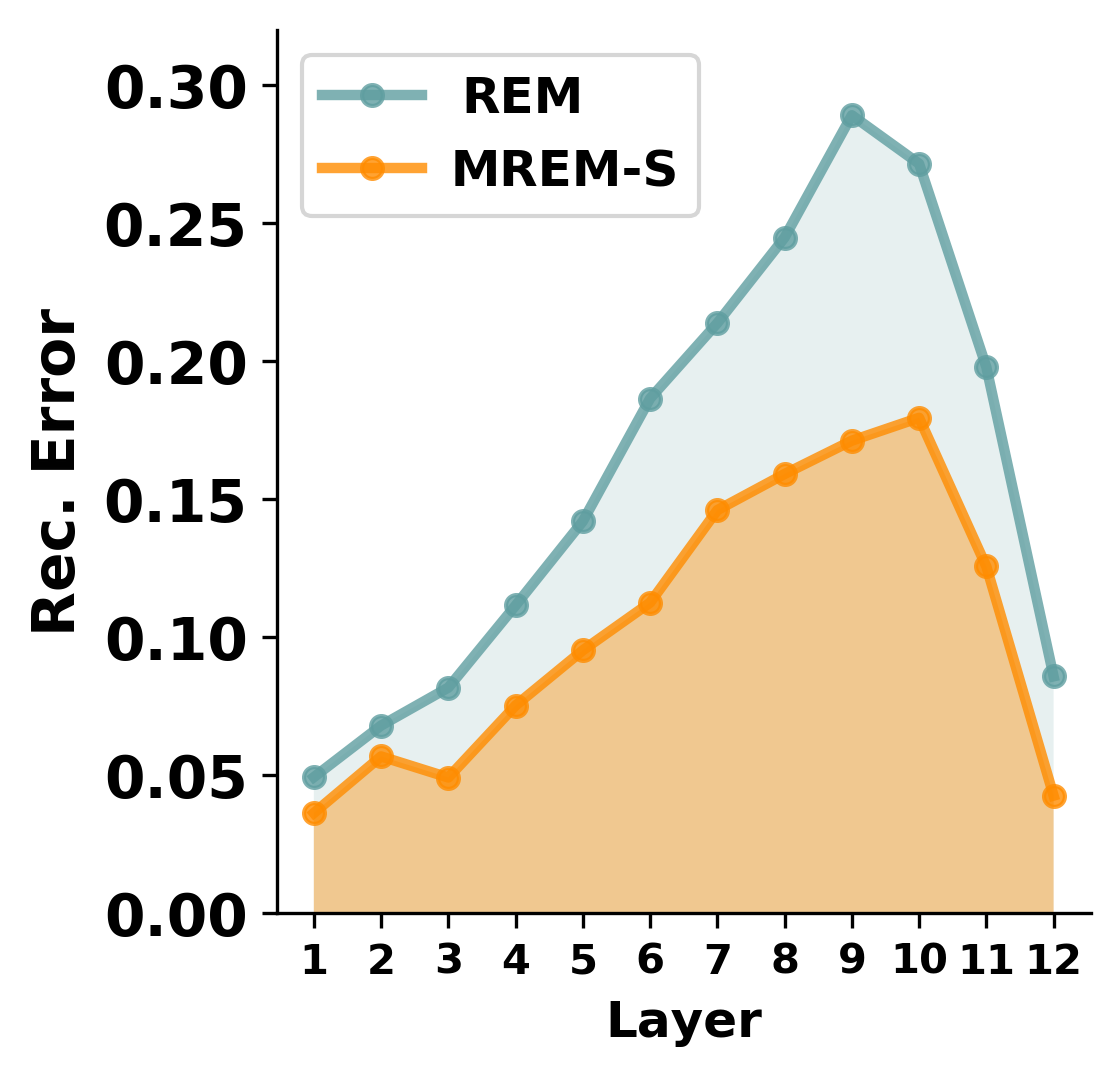}
	    \label{fig:error_propagate_a4}
	}
	\vspace{-1.5ex}
    \caption{Discussions on the proposed MREM approach. In (a) and (b), the solid line and shaded area denote the averaged results and standard deviation of a ``W2-E2-A4'' quantized BERT-base model from 10 different seeds. (c) and (d) visualize the propagation of reconstruction error on ``W2-E2-A8'' and ``W2-E2-A4'' quantized BERT-base model, respectively.}
    \label{fig:ablation}
\end{figure*}

\subsubsection{Teacher Forcing}
\label{exp:teacher_forcing}
We now study how teacher forcing benefits MREM-P with different numbers of training steps, and results are listed in Table~\ref{tab:teacher_forcing}.
It can be found that teacher forcing can bring consistent improvement for both BERT-base and BERT-large models. Moreover, the gain of teacher forcing is more significant with fewer training steps or lower quantization bit-width, i.e., $\textbf{3.4}\%\uparrow$ and $\textbf{2.8}\%\uparrow$ on the ``W2-E2-A4'' quantized BERT-base and BERT-large respectively under $250$ steps. This matches our intuition that fewer training steps or higher compression ratio give larger reconstruction error, when the clean input from the full-precision module can benefit more the quantized module.
As further increasing the training steps brings only marginal improvement and diminishes the effect of teacher forcing, we by default set the training steps to $2,000$.

Additionally, we also plot training loss curves of the four modules under 250 and 2,000 training steps in Figure~\ref{fig:parallel_loss_curve}. 
We find that: 1) the loss curves with teacher forcing are apparently lower, especially when trained with fewer steps, which matches the observations in Table~\ref{tab:teacher_forcing}; 2) the loss curves of late modules are usually lower than the earlier ones, indicating that modules closer to the output benefit more from teacher forcing. This matches the intuition that the late modules have more errors accumulated to correct.
The red areas in Figure~\ref{fig:parallel_loss_curve} show that  $40\%$ of the total iterations are used for teacher forcing. We also try tuning it within $[20\%, 80\%]$, and does not observe large difference in the final performance. We thus choose $40\%$ by default.

\subsubsection{Further Comparison with REM}
\label{sec:longer_rem}

Here we provide further discussions with REM on the training efficiency.
Note that both REM and MREM-S follow the sequential training procedure, where the output from the previous objective is cached for the next objective. 
However, as there are many matrix multiplications in each Transformer layer, it can be time-consuming for REM to repeat this procedure recursively.
According to results in Section~\ref{sec:main_results}, while REM and MREM take roughly the same amount of time,
REM is only iterated for $250$ steps on MNLI and $500$ steps on SQuAD, while MREM takes $2,000$ steps and $4,000$ steps respectively.


We also provide results when REM takes the same amount of training steps with MREM-S in Table~\ref{tab:REM_MREM}.
It can be found that even with $2,000$ iterations, REM is still inferior to MREM-S across all quantization bit-widths. 
Meanwhile, REM nearly takes around $9\times$ more training time than MREM.
Therefore, the module-wise granularity in MREM not only improves the quantization performance with more layer-wise dependencies considered, but also makes the training pipeline efficient with fewer stages to cache intermediate results.



\subsubsection{Number of Modules and Memory Overhead}
\label{sec:num_module}
We verify the effect of model partition on the final quantized performance, as well as their corresponding memory consumption. 
According to Figure~\ref{fig:num_partition}, by varying
the number of modules within $\{1,2,3,4,6\}$, it can be found that fewer model partitions give slightly better performance, as layer-wise dependencies can be better incorporated for reconstruction error minimization.
However, this also comes with more running memory, i.e., $\{11.9, 6.7, 4.7, 3.7, 2.7\}$ GB for these partitions correspondingly. The decrease of memory also diminishes with fewer partitions.
Therefore. as a trade-off, we partition the model into 4 modules by default.

\subsubsection{Size of Calibration Data}
The size of calibration data directly relates to the security and privacy issues in post-training quantization. 
To learn its effects, we vary the calibration data size $|\tilde{\mathcal{D}}|$ within $\{32, 64, 128, 512, 1024, 2048, 4096, 8192\}$, and list the results of REM, MREM-S and MREM-P.
From Figure~\ref{fig:data_size}, it can be found that while REM is ahead of MREM-S/P with fewer than $128$ training samples, the accuracy of REM rises slowly and saturates at around $60\%$ afterwards.
We hypothesize that the simple training objective in REM can hardly hold more training instances for optimization.
MREM-S/P, on the other hand, can better exploit larger calibration data size, since the module-wise granularity admits higher flexibility for the optimization.
As we find the diminishing gain to increase the training size after $4,096$ samples, we by default take $4,096$ samples.

\subsubsection{Reconstruction Error Propagation}
We visualize the propagation of reconstruction error for both ``W2-E2-A8'' and ``W2-E2-A4'' quantized BERT-base models in Figure~\ref{fig:error_propagate_a8} and Figure~\ref{fig:error_propagate_a4} respectively. It can be observed that our MREM achieves both lower values and slower rising rates of the reconstruction error than REM across all layers, which verifies the advantage of module-wise granularity to minimize the reconstruction error.
Interestingly, while the reconstruction error generally gets enlarged layer-wisely in the first ten layers, it begins to decrease afterwards. 
We speculate this is due to the effect of the classification head that encourages concentrated hidden representations for the task.

\begin{table}
\resizebox{0.48\textwidth}{!}{
\begin{tabular}{cc|cc|cc}
	\hline\hline
	    \multirow{3}{*}{\textbf{\tabincell{c}{\#Bits\\(W-E-A)}}} &
	    \multirow{3}{*}{\textbf{Methods}} & \multicolumn{2}{c}{\textbf{w/o PCQ}} & \multicolumn{2}{c}{\textbf{w/ PCQ}} \\
	    \cline{3-4}\cline{5-6}
	    &  & \textbf{\tabincell{c}{Acc\\m(\%)}} & \textbf{\tabincell{c}{Acc\\mm(\%)}} &  \textbf{\tabincell{c}{Acc\\m(\%)}} & \textbf{\tabincell{c}{Acc\\mm(\%)}} \\\hline
	    \multirow{2}{*}{4-4-8} & REM & $73.3_{\pm 0.3}$ & $74.9_{\pm 0.2}$ & $75.9_{\pm 0.3}$ & $77.4_{\pm 0.2}$ \\
	     & MREM & $83.5_{\pm 0.1}$ & $83.9_{\pm 0.2}$ & $83.6_{\pm 0.1}$ & $84.0_{\pm 0.1}$ \\	    
         \hline
	    \multirow{2}{*}{2-2-8} & REM & $71.6_{\pm 0.4}$ & $73.4_{\pm 0.4}$ & $74.1_{\pm 0.5}$ & $75.6_{\pm 0.5}$ \\
	     & MREM & $82.7_{\pm 0.2}$ & $82.7_{\pm 0.2}$ & $82.8_{\pm 0.1}$ & $82.9_{\pm 0.1}$ \\\hline
	    \multirow{2}{*}{2-2-4} & REM & $58.3_{\pm 0.5}$ & $60.6_{\pm 0.6}$ & $59.3_{\pm 0.4}$ & $62.0_{\pm 0.4}$ \\
	     & MREM & $81.1_{\pm 0.2}$ & $81.5_{\pm 0.2}$ & $81.1_{\pm 0.2}$ & $81.5_{\pm 0.3}$ \\	     
         \hline\hline
	\end{tabular}}
	\caption{Comparison of BERT-base results with and without per-channel quantization~(PCQ) on MNLI.}
	\label{tab:pcq}
\end{table}

\subsubsection{Per-channel Quantization}
Per-channel Quantization~(PCQ) is prevalent in the post-training quantization of convolution neural networks~\cite{nahshan2019loss,nagel2020up,hubara2020improving}. To learn its effect in PLMs, PCQ assigns different quantization step-sizes at each output dimension of the linear layer, which is also known as row-wise quantization in~\cite{zhang2020ternarybert}.
The PCQ results of REM and MREM are shown in Table~\ref{tab:pcq}.
It can be found that while PCQ improves REM by around $1.0\%$ to $2.5\%$, the gain is very incremental on MREM.
We hypothesize that more training steps of MREM can better adjust the quantization distribution for PLMs.
Our results are also similar to the findings in ~\cite{zhang2020ternarybert}, where the row-wise quantization brings little improvement.
As PCQ also requires to store more full-precision step sizes, we do not employ PCQ by default.


\section{Conclusion}
\label{sec:conclusion}

In this paper, we study post-training quantization for pre-trained language models. 
We show that existing quantization-aware training solutions suffer from slow training, 
huge memory overhead, and data privacy issues when accessing the full training set.
To mitigate these issues, we propose module-wise reconstruction error minimization, 
an efficient solution to quantize PLMs. MREM can be conducted either sequentially or in parallel, 
where the parallel training can achieve the speedup close to the theoretical limit without apparent performance degradation. 
Experimental results show that the proposed solution greatly improves the performance. Meanwhile, it significantly reduces the training time and memory overhead with only thousands of training instances.

There are several promising directions to explore in the future: 1) We can scale the proposed approach to larger PLMs, and thus more models can benefit from post-training quantization; 2) The proposed parallel strategy can be applied to warm up the pre-training of PLMs, such that the overall pre-training cost can be reduced; 3) While the current parallel strategy conducts local training separately, it would be interesting to cache module-wise gradients in some queues for the backward pass so that there is less discrepancy with conventional end-to-end training.

\bibliographystyle{IEEEtran}
\bibliography{ref}

\end{document}